\documentclass{article}

    \PassOptionsToPackage{numbers, compress}{natbib}

 \usepackage[main, final]{neurips_2026}

\usepackage[utf8]{inputenc} %
\usepackage[T1]{fontenc}    %
\usepackage{hyperref}       %
\usepackage{url}            %
\usepackage{booktabs}       %
\usepackage{amsfonts}       %
\usepackage{nicefrac}       %
\usepackage{microtype}      %
\usepackage[table]{xcolor}         %

\usepackage{float}
\usepackage{graphicx}
\usepackage{amsmath}
\usepackage{kotex}
\usepackage{adjustbox}
\usepackage{enumitem} %
\usepackage{makecell}
\usepackage{multirow}
\usepackage{hhline}
\usepackage{wrapfig}

\title{SPA-Q: Structure-Preserving Adaptive Post-Training Quantization for Monocular Depth Estimation}

\author{
  Jaemin Choi$^{1}$\thanks{Equal contribution.} \quad
  Jincheol Yang$^{1}$\footnotemark[1] \quad
  Nahyun Lim$^{1}$\footnotemark[1] \quad
  Yun-Seong Jeong$^{1}$\footnotemark[1] \\
  \textbf{Matti Zinke$^{2}$\footnotemark[1] \quad
  Hyunwoo Yu$^{1}$\footnotemark[1] \quad
  Suk-Ju Kang$^{1}$\thanks{Corresponding author.}} \\[3pt]
  $^{1}$Department of Electronic Engineering, Sogang University \\
  $^{2}$Department of Computer Science, Sogang University \quad \\
  {\small\texttt{\{jam0225, yjc3232, julialim5, jus8077, mattizinke, hyunwoo137, sjkang\}@sogang.ac.kr}}
}

\begin{document}

\maketitle
\begin{abstract}

Monocular depth estimation (MDE) has advanced rapidly with the emergence of foundation models such as Depth Anything. 
Their transformer-based architectures provide strong generalization across diverse scenes and domains, but also incur high computational and memory cost, making efficient deployment challenging.
Post-Training Quantization (PTQ) provides an efficient and practical solution for model compression, yet low-bit PTQ remains challenging for MDE models.
When applying PTQ to Depth Anything, we identify two key challenges: (i) independently minimizing the quantization error of the query and key projections fails to preserve the attention maps induced by their interaction, and (ii) quantization errors accumulate across layers, resulting in intermediate feature distribution shifts.
To address these issues, we propose SPA-Q, a Structure-Preserving Adaptive PTQ framework for MDE, consisting of two main components.
First, Attention-Preserving Calibration (APC) determines query and key quantization parameters by matching the full-precision attention distribution.
Second, Channel-Wise Distribution Alignment (CWDA) learns channel-wise affine transformations to mitigate quantization-induced distribution shifts, and the learned parameters are absorbed into the weights after training.
Experimental results show that SPA-Q consistently outperforms existing PTQ methods under 4-bit quantization, achieving an average 25.9\% reduction in AbsRel and a 17.3\% improvement in $\delta_1$ across NYUv2 and KITTI datasets.
\end{abstract}
    
\section{Introduction}

Monocular depth estimation (MDE) is a fundamental task in computer vision, with applications in robotics \cite{survey,fastdepth}, autonomous driving \cite{autonomous1,autonomous2}, and 3D scene understanding \cite{nerf, gaussian, TII}. 
Recent advances in Vision Transformers (ViTs) \cite{vit, deit, DINOv1, DINOv2, embedding} have significantly improved MDE, leading to strong foundation models such as MiDaS \cite{MiDaS} and Depth Anything \cite{DAv1, DAv2}.
Depth Anything achieves strong zero-shot generalization through large-scale training with labeled and pseudo-labeled data, while leveraging multi-scale intermediate features for dense depth prediction. 
However, its large model size and high computational cost hinder real-world deployment in resource-constrained environments.

Model quantization \cite{quant_survey, white, integer, low, adaround, FQ-ViT} is an effective approach for reducing the memory and computational cost of deep neural networks. 
Existing quantization methods can be broadly divided into quantization-aware training (QAT) \cite{LSQ, LSQ+, PACT} and post-training quantization (PTQ) \cite{BRECQ, QDROP, RepQ-ViT, erq}. 
QAT incorporates quantization effects during training and often achieves higher accuracy, but it requires additional training and access to training data. 
In contrast, PTQ applies quantization to a pre-trained model without full retraining, making it suitable for practical deployment.
Although PTQ has been widely studied for recognition models, its application to MDE remains underexplored.
Unlike recognition tasks, depth estimation requires preserving fine-grained geometric and structural information in intermediate features, making it more sensitive to quantization-induced distortions.

In our study, we identify two key limitations of low-bit PTQ for transformer-based MDE models. 
\textbf{(i) Independent quantization error minimization for query and key projections is not well aligned with preserving the resulting attention distribution.} 
Existing PTQ methods \cite{BRECQ, QDROP, RepQ-ViT, erq} typically calibrate each quantizer by minimizing the discrepancy between full-precision and quantized outputs of each projection.
However, in self-attention, the attention map is determined by the query-key interaction passed through the softmax operation, rather than by the fidelity of each projection. 
Therefore, independently minimizing the quantization error of the query and key projections does not guarantee preservation of the resulting attention map. 
This issue is especially critical for MDE, as distorted attention maps can corrupt features aggregated by the decoder, thereby degrading scene geometry and object boundaries in depth predictions.
\textbf{(ii) Quantization errors accumulate across layers and induce distribution shifts in intermediate features.}
Since the DPT decoder fuses features from multiple encoder stages, even small shifts in feature distributions can propagate through the decoder and degrade depth prediction. 
In practice, low-bit quantization induces noticeable distribution shifts in intermediate features, making them increasingly mismatched with full-precision counterparts.

To address these limitations, we propose \textbf{SPA-Q}, a novel \textbf{S}tructure-\textbf{P}reserving \textbf{A}daptive PTQ framework for transformer-based MDE models.
SPA-Q is designed for architectures whose decoders aggregate multi-stage encoder features to produce dense depth predictions.
It consists of two complementary components.
\textbf{Attention-Preserving Calibration (APC)} determines the quantization parameters of the query and key projections by directly preserving the full-precision attention distribution instead of minimizing projection errors independently.
In particular, APC minimizes the KL divergence between the full-precision and quantized attention distributions, thereby better capturing their discrepancy after softmax.
\textbf{Channel-Wise Distribution Alignment (CWDA)} addresses distribution shifts in intermediate features through learnable channel-wise affine transformations, which are optimized with feature-level and depth-level objectives in an end-to-end manner.
The learned affine parameters are then fused into the weights, introducing no additional inference overhead.
Our contributions are summarized as follows:
\begin{itemize}
\item We identify two limitations of low-bit PTQ for transformer-based MDE models. First, independent quantization error minimization for query-key projections fails to preserve the resulting attention map. Second, accumulated quantization errors induce distribution shift in intermediate features. 

\item We propose SPA-Q, a Structure-Preserving Adaptive PTQ method for MDE. SPA-Q introduces Attention-Preserving Calibration (APC), which calibrates query and key quantizers with an attention-level objective, and Channel-Wise Distribution Alignment (CWDA), a channel-wise alignment scheme for correcting intermediate feature distribution shifts. 

\item Extensive experiments on relative depth estimation benchmarks show that SPA-Q outperforms existing PTQ methods under 4-bit quantization across all evaluated settings.
\end{itemize}

\section{Related Work}

\subsection{Monocular Depth Estimation Models}

MDE \cite{bts, zoedepth} refers to the task of predicting a dense depth map from a single image, where each pixel reflects the 3D structure of the scene. 
Recent advances driven by large-scale data and transformer-based architectures have significantly improved depth prediction, leading to models such as MiDaS \cite{MiDaS} and Depth Anything \cite{DAv1,DAv2}. 
MiDaS improves generalization by training on multiple datasets with varying scales and distributions, reducing dependence on a single dataset. 
Building on this, Depth Anything adopts a DINOv2 \cite{DINOv2}-pretrained ViT encoder. Depth Anything v1 leverages both labeled and unlabeled data, while v2 further improves performance by incorporating large-scale synthetic and pseudo-labeled data. 
Both MiDaS and Depth Anything are typically built on the DPT \cite{DPT} architecture, which aggregates multi-scale features using a transformer encoder and reconstructs them with a multi-scale decoder. 
We focus on Depth Anything v1 and v2 for evaluation, as they represent state-of-the-art MDE models with strong generalization performance.

\subsection{Post-Training Quantization}

PTQ \cite{quant_survey, white, integer} has been widely studied as an efficient approach for compressing neural networks without requiring full retraining.
Most existing PTQ methods are designed for recognition tasks such as image classification. BRECQ \cite{BRECQ} minimizes quantization error via block-wise reconstruction using a second-order approximation. 
By randomly dropping activation quantization during reconstruction, QDrop \cite{QDROP} mitigates overfitting and improves generalization.
RepQ-ViT \cite{RepQ-ViT} further reduces inter-channel variation through scale reparameterization, enabling channel-wise calibration with tensor-wise inference. 
However, these methods have been primarily designed for and evaluated on classification tasks \cite{vit, deit, swin}, and do not adequately address dense prediction tasks.
In MDE, predictions rely on multi-scale intermediate features, making the task more sensitive to quantization errors. 
To address this limitation, QSCA \cite{QSCA} introduces lightweight auxiliary modules for residual error compensation. 
Despite its effectiveness, this approach incurs additional parameters and computational overhead, which is undesirable for efficient deployment.  
In this work, we propose a PTQ method for MDE that preserves intermediate features required for dense depth prediction by maintaining attention maps and mitigating feature distribution shifts.

\section{Method}

\subsection{Preliminaries}

\textbf{Uniform quantizer.}
A uniform quantizer \cite{quant_survey, white, integer} maps full-precision values to evenly spaced discrete levels.
Given an input \(x\) and bit-width \(b\), the quantized integer \(\bar{x}\) and the dequantized value \(\hat{x}\) are computed as:
\begin{equation}
\label{eqn:uniformquant}
\bar{x} = \text{clip}\left( \left\lfloor \frac{x}{s} \right\rceil + z, 0, 2^b - 1 \right), \quad
\hat{x}=s \cdot (\bar{x}-z),
\end{equation}
where the step size \(s\) and zero-point \(z\) are
\begin{equation}
\label{eqn:scale_zeropoint}
s = \frac{\max(x) - \min(x)}{2^b - 1}, \quad
z = \left\lfloor \frac{-\min(x)}{s} \right\rceil ,
\end{equation}
and \(\lfloor \cdot \rceil\) denotes rounding to the nearest integer, while \(\text{clip}(\cdot, 0, 2^b-1)\) constrains the quantized integer to the valid range \([0, 2^b-1]\).

\textbf{Log2 quantizer.}
As a basic form of logarithmic quantizer, the log$2$ quantizer \cite{FQ-ViT} maps inputs to quantization levels defined on a base-2 logarithmic scale.
Compared to uniform quantization, it allocates more quantization levels to small values, making it better suited for highly skewed or long-tailed distributions such as post-Softmax activations. 
Given an input \(x\) and step size \(s\), the quantized value \(\hat{x}\) is computed as:
\begin{equation}
\bar{x} = \text{clip} \left( \left\lfloor -\log_{2} \frac{x}{s} \right\rceil, \, 0, \, 2^b - 1 \right), \quad
\hat{x}=s \cdot 2^{-\bar{x}}.
\label{eq:log2}
\end{equation}
The log$2$ quantizer defines quantization levels on an exponential scale, making it more effective for values concentrated near zero. Accordingly, we adopt the log$2$ quantizer for post-Softmax activations.

\subsection{Motivation}
To effectively quantize transformer-based MDE models, we analyze how low-bit quantization degrades representations used for dense depth prediction.
In particular, we examine two sources of distortion: attention maps resulting from query-key interaction and intermediate feature distributions.

\textbf{Mismatch between local quantization error minimization and attention preservation.}
Existing PTQ methods typically calibrate the query and key projections by minimizing local quantization error. 
Let \(X_Q \in \mathbb{R}^{N_q \times D}\) and \(X_K \in \mathbb{R}^{N_k \times D}\) denote the input features to the query and key projections, respectively, where \(N_q\) and \(N_k\) are the numbers of query and key tokens, and \(D\) is the input channel dimension. 
The query, key, and attention map are then given by:
\begin{equation}
\mathbf{Q} = X_Q W_Q, \qquad
\mathbf{K} = X_K W_K, \qquad
\mathbf{A} = \mathrm{softmax}\left(\frac{\mathbf{Q}\mathbf{K}^{\top}}{\sqrt{d_h}}\right),
\end{equation}
where \(W_Q, W_K \in \mathbb{R}^{D \times d_h}\) are the corresponding projection weights, and \(d_h\) denotes the head dimension.
This formulation shows that the attention map is determined by the query-key interaction through 
\begin{figure}[h]
    \centering
    \includegraphics[width=0.98\textwidth]{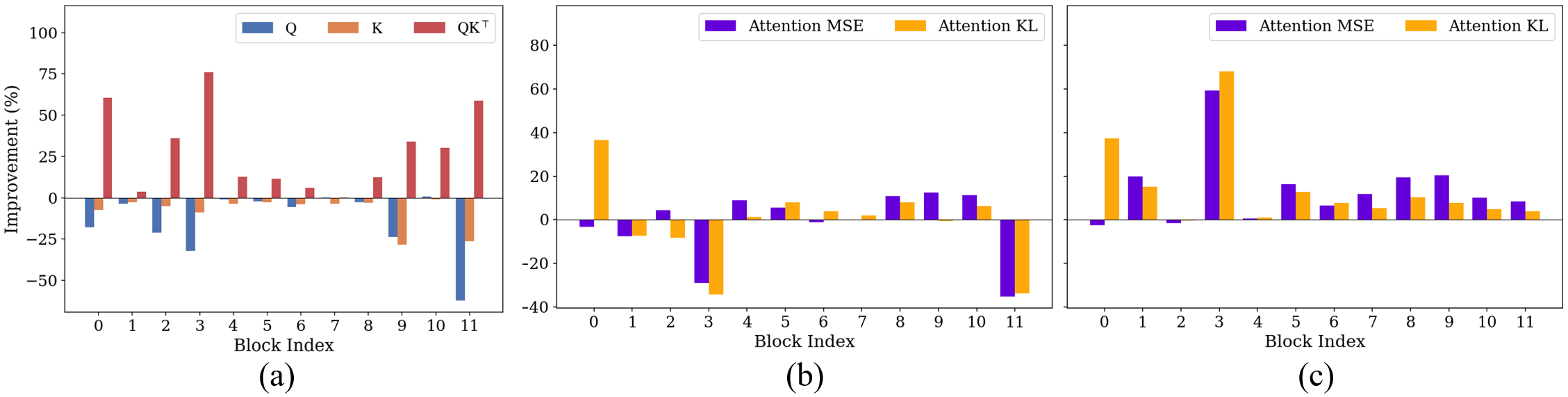}
    \caption{Comparison of calibration objectives for query-key quantization, where all improvements are measured relative to conventional local quantization error minimization. (a) MSE improvement from logits-level calibration. (b) Attention-level metric improvement from logits-level calibration. (c) Attention-level metric improvement from the proposed APC.}
    \label{fig:fig_apc}    
\end{figure}
\begin{figure}[h]
    \centering
    \includegraphics[width=0.98\textwidth]{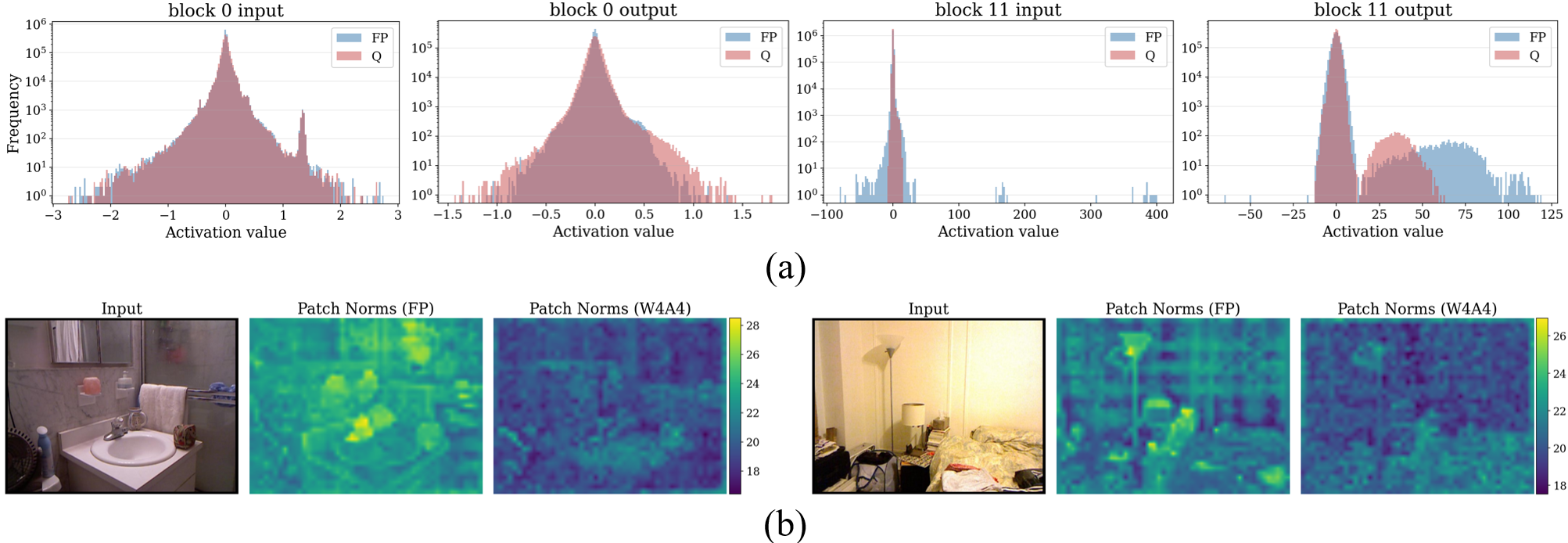}
    \caption{Quantization-induced distortion in intermediate features under 4-bit quantization (W4A4). (a) Distribution shift in intermediate features caused by accumulated quantization errors across layers. (b) Distortion of intermediate feature maps under W4A4, as illustrated by comparisons of patch norms with the full-precision model.}
    \label{fig:fig_cwda}
\end{figure}
the pre-softmax logits \(\mathbf{Q}\mathbf{K}^{\top}/\sqrt{d_h}\), rather than by the fidelity of each projection alone.
Nevertheless, conventional PTQ minimizes the local quantization error of the two projections independently:
\begin{equation}
\theta_{Q}^{\star} = \arg\min_{\theta_{Q}} \left\| \mathbf{Q} - \hat{\mathbf{Q}}_{\theta_{Q}} \right\|_F^2,
\qquad
\theta_{K}^{\star} = \arg\min_{\theta_{K}} \left\| \mathbf{K} - \hat{\mathbf{K}}_{\theta_{K}} \right\|_F^2,
\label{eq:local_qk}
\end{equation}
where $\theta_Q=\{s_Q,z_Q\}$ and $\theta_K=\{s_K,z_K\}$ denote the corresponding quantization parameter sets. This objective implicitly assumes that preserving $\mathbf{Q}$ and $\mathbf{K}$ individually is sufficient for preserving the resulting attention map, even though the attention map is produced by their interaction.

To verify this mismatch empirically, we replace the local quantization error minimization in Eq.~\eqref{eq:local_qk} with a calibration objective defined on the pre-softmax attention logits:
\begin{equation}
\theta_{Q}^{\star}, \theta_{K}^{\star}
= \arg\min_{\theta_{Q},\theta_{K}}
\left\| \frac{\mathbf{Q}\mathbf{K}^{\top}}{\sqrt{d_h}}
- \frac{\hat{\mathbf{Q}}_{\theta_{Q}}\hat{\mathbf{K}}_{\theta_{K}}^{\top}}{\sqrt{d_h}} \right\|_F^2.
\label{eq:joint_qk}
\end{equation}
As shown in Figure~\ref{fig:fig_apc} (a), although this logits-level objective often degrades the MSE of $\mathbf{Q}$ and $\mathbf{K}$ individually, it consistently improves the MSE of the attention logits $\mathbf{Q}\mathbf{K}^{\top}$. 
This observation reveals a fundamental mismatch between local quantization error minimization and query-key interaction: optimizing the two projections individually does not necessarily preserve their interaction. 
However, reducing the MSE of the attention logits does not consistently improve the attention map under MSE and KL divergence, as shown in Figure~\ref{fig:fig_apc} (b).
This gap arises from the nonlinearity of the softmax. The attention logits are unnormalized real-valued scores, whereas the attention map is a normalized probability distribution after softmax. 
Consequently, minimizing the Euclidean error of the attention logits is not equivalent to minimizing the discrepancy between the corresponding attention maps.

\begin{figure}[h]
    \centering
    \includegraphics[width=0.95\textwidth]{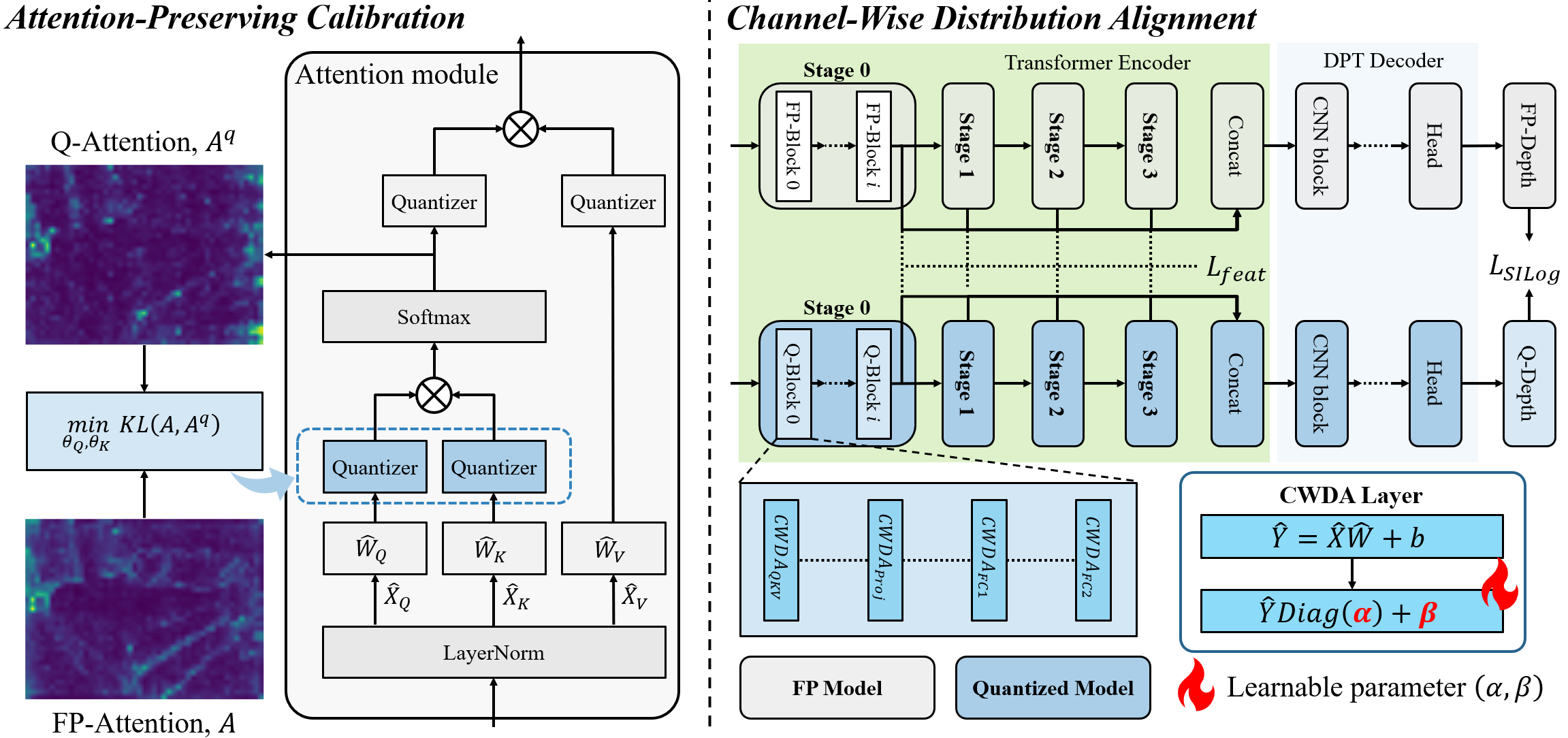}
    \caption{
Overview of the SPA-Q framework.
SPA-Q first applies Attention-Preserving Calibration (APC) to determine query and key quantization parameters by minimizing the row-wise KL divergence between full-precision and quantized attention maps.
It then applies Channel-Wise Distribution Alignment (CWDA), where learnable affine parameters align quantized intermediate features with their full-precision counterparts using stage-wise feature maps from the transformer encoder and the depth prediction from the decoder.
After optimization, the learned affine parameters are fused into the weights, introducing no additional inference overhead.
}
    \label{fig:fig_framework}
    
\end{figure}

\textbf{Feature distortion from accumulated quantization errors.}
Low-bit quantization progressively shifts intermediate feature distributions away from their full-precision counterparts, as illustrated in Figure~\ref{fig:fig_cwda} (a). 
This distortion arises because the quantized output of each layer is fed into the next layer, causing quantization errors to accumulate across layers. 
Under tensor-wise quantization, this accumulation can become more severe because a single quantization scale is shared across all channels, making the quantization error more sensitive to inter-channel variation \cite{RepQ-ViT, dopq, uq-vit}. 
As a result, the distribution shift can be further amplified, distorting not only feature statistics but also the corresponding feature maps, as shown in Figure~\ref{fig:fig_cwda} (b). 
This issue is especially critical in transformer-based MDE models, where the decoder relies on stage-wise features from multiple encoder blocks. 
Once these distorted features are propagated to the decoder, the quality of depth prediction deteriorates.
\label{sec:motivation}

\subsection{Attention-Preserving Calibration}

As discussed in Sec.~\ref{sec:motivation}, local quantization error minimization for query and key projections is not well aligned with preserving the resulting attention map, and logits-level calibration is still insufficient for matching the attention distribution after softmax.
Based on this observation, we propose \textbf{Attention-Preserving Calibration (APC)}, which determines the query and key quantization parameters by directly minimizing the discrepancy between the full-precision and quantized attention distributions.
Unlike objectives defined on individual projections or pre-softmax attention logits, APC defines the calibration objective on the attention map.
Since each row of the attention map forms a normalized probability distribution over tokens, KL divergence is more appropriate than Euclidean error for measuring discrepancies in probability mass and capturing distortions in the attention distribution.
Accordingly, APC minimizes the row-wise KL divergence between the full-precision attention map \(\mathbf{A}\) and the quantized attention map \(\mathbf{A}^{q}\):
\begin{equation}
\theta_{Q}^{\star}, \theta_{K}^{\star}
=
\arg\min\limits_{\theta_{Q},\theta_{K}}
\frac{1}{N_q}\sum_{i=1}^{N_q}
D_{\mathrm{KL}}\!\left(
\mathbf{A}_{i,:}
\,\middle\|\,
\mathbf{A}^{q}_{i,:}
\right),
\label{eq:apc_rowwise}
\end{equation}
where \(\mathbf{A}^{q}=\mathrm{softmax}(
\hat{\mathbf{Q}}_{\theta_Q}\hat{\mathbf{K}}_{\theta_K}^{\top}/\sqrt{d_h})\). 
As shown in Figure~\ref{fig:fig_apc} (c), the proposed APC objective consistently reduces both the MSE and KL divergence between the full-precision and quantized attention maps across most blocks. 
This demonstrates that attention-level calibration better aligns the quantized attention map with its full-precision counterpart than projection-wise or logits-level objectives. 
As a result, APC better maintains the structural information used to construct intermediate features for depth prediction.

\subsection{Channel-Wise Distribution Alignment}

To address the intermediate feature distribution shifts identified in Sec.~\ref{sec:motivation}, we propose \textbf{Channel-Wise Distribution Alignment (CWDA)}, which aligns dequantized output activations with their full-precision counterparts through learnable channel-wise affine transformations.
CWDA can be applied to output activations from both fully connected and convolutional layers.
For notational simplicity, we represent the dequantized output activation as \(\hat{\mathbf{Y}} \in \mathbb{R}^{N \times D}\), where \(N\) denotes the number of tokens or flattened spatial positions, and \(D\) denotes the number of channels.
The aligned output \(\hat{\mathbf{Y}}^{\star}\) is then given by:
\begin{equation}
\hat{\mathbf{Y}}^{\star}_{:,d}=\alpha_d \hat{\mathbf{Y}}_{:,d} + \beta_d,\quad d = 1, \dots, D,
\label{eq:cwda_affine}
\end{equation}
where \(\alpha_d\) and \(\beta_d\) are learnable scaling and shifting parameters for the \(d\)-th channel, optimized on training set.
Defining \(\boldsymbol{\alpha}=[\alpha_1,\dots,\alpha_D]^\top\) and \(\boldsymbol{\beta}=[\beta_1,\dots,\beta_D]^\top\), Eq.~\eqref{eq:cwda_affine} becomes
\begin{equation}
\hat{\mathbf{Y}}^{\star}=\hat{\mathbf{Y}}\,\operatorname{Diag}(\boldsymbol{\alpha})+\boldsymbol{\beta}.
\label{eq:cwda_matrix}
\end{equation}
Since the affine transformation is applied independently to each output channel, it can be absorbed into the weight and bias of the preceding transformation:
\begin{equation}
\hat{\mathbf{Y}}^{\star}
=
\hat{\mathbf{X}}\hat{\mathbf{W}}\operatorname{Diag}(\boldsymbol{\alpha})
+
\mathbf{b}\odot\boldsymbol{\alpha}
+
\boldsymbol{\beta}
=
\hat{\mathbf{X}}\hat{\mathbf{W}}^{\star}
+
\mathbf{b}^{\star},
\label{eq:cwda_absorb}
\end{equation}
where \(\hat{\mathbf{W}}^{\star}=\hat{\mathbf{W}}\operatorname{Diag}(\boldsymbol{\alpha})\) and \(\mathbf{b}^{\star}=\mathbf{b}\odot\boldsymbol{\alpha}+\boldsymbol{\beta}\). Therefore, CWDA introduces no additional inference overhead after training.

\textbf{End-to-end optimization.}
The DINOv2 \cite{DINOv2} encoder and DPT \cite{DPT} decoder jointly determine the final depth prediction in MDE. 
Quantization-induced distortion in intermediate features propagates through the network and ultimately affects the predicted depth map.
To account for both representation-level and output-level effects, we optimize all CWDA parameters in an end-to-end manner using feature-level and depth-level objectives.
To preserve the stage-wise features delivered to the decoder, we minimize the $\ell_1$ distance between the full-precision and quantized stage features:
\begin{equation}
    \mathcal{L}_{\text{feat}} 
    = \frac{1}{|\mathcal{S}|} \sum_{s \in \mathcal{S}} 
      \left\| \mathbf{F}_{s} - \hat{\mathbf{F}}_{s} \right\|_1,
\end{equation}
where $\mathcal{S}$ denotes the set of stage indices whose outputs are fed to the decoder, and $\mathbf{F}_{s},\, \hat{\mathbf{F}}_{s} \in \mathbb{R}^{N \times C}$ denote the full-precision and quantized feature maps at stage $s$, respectively. 
To preserve the output depth structure, we minimize the SILog loss \cite{MiDaS} between full-precision and quantized depth maps:
\begin{equation}
    \mathcal{L}_{\text{SILog}}
    =
    \frac{1}{n}\sum_{i=1}^{n} z_i^2
    -
    \frac{1}{n^2}\left(\sum_{i=1}^{n} z_i\right)^2,
    \qquad
    z_i = \log \hat{d}_i - \log d_i,
\end{equation}
where $d_i$ and $\hat{d}_i$ denote the full-precision and quantized depth maps at pixel $i$, respectively, and $n$ is the number of valid pixels. The final objective is defined as:
\begin{equation}
    \mathcal{L}
    =
    \mathcal{L}_{\text{feat}}
    +
    \lambda_{\text{SILog}} \mathcal{L}_{\text{SILog}},
    \label{eq:joint_loss}
\end{equation}
which is used to jointly optimize all CWDA parameters in the encoder and decoder.

\section{Experiments}

\subsection{Experimental Setup}

\paragraph{Models, datasets, and evaluation.}

We use Depth Anything v1 \cite{DAv1} and v2 \cite{DAv2}, built on the DINOv2 \cite{DINOv2} encoder with ViT-S and ViT-B backbones, as representative foundation models for monocular depth estimation.
We evaluate SPA-Q on five MDE benchmarks: NYUv2 \cite{nyu}, KITTI \cite{kitti}, ETH3D \cite{eth3d}, DIODE \cite{diode}, and Sintel \cite{sintel}. 
These datasets correspond to the official zero-shot relative depth evaluation benchmarks used in Depth Anything. For evaluation, we report standard depth estimation metrics, including the absolute relative error (AbsRel) and threshold accuracy \(\delta_1\). 
AbsRel measures the mean relative difference between predicted and reference depth values, while \(\delta_1\) measures the percentage of predictions satisfying a threshold ratio of \(1.25\). 

\begin{table}
\centering
\caption{Comparison of PTQ methods on NYUv2 \cite{nyu}, KITTI \cite{kitti}, ETH3D \cite{eth3d}, DIODE \cite{diode}, and Sintel \cite{sintel} for zero-shot relative depth estimation under 4-bit quantization. \textit{E.} denotes the encoder backbone used in the MDE architectures.}

\setlength{\tabcolsep}{10pt}
\renewcommand{\arraystretch}{1.5}
\resizebox{\textwidth}{!}{
\begin{tabular}{c l c c c c c c c c c}
\Xhline{1.4pt}
\multirow{3}{*}{\textbf{Dataset}} & \multirow{3}{*}{\textbf{Method}} & \multirow{3}{*}{\textbf{W/A}}
& \multicolumn{4}{c}{\textbf{Depth Anything v1}} 
& \multicolumn{4}{c}{\textbf{Depth Anything v2}} \\
\cmidrule(lr){4-7} \cmidrule(lr){8-11}
& & 
& \multicolumn{2}{c}{\textbf{\textit{E.} ViT-S}} 
& \multicolumn{2}{c}{\textbf{\textit{E.} ViT-B}} 
& \multicolumn{2}{c}{\textbf{\textit{E.} ViT-S}} 
& \multicolumn{2}{c}{\textbf{\textit{E.} ViT-B}} \\
& & 
& AbsRel $\downarrow$ & $\delta_1 \uparrow$
& AbsRel $\downarrow$ & $\delta_1 \uparrow$
& AbsRel $\downarrow$ & $\delta_1 \uparrow$
& AbsRel $\downarrow$ & $\delta_1 \uparrow$ \\
\midrule
\multirow{6}{*}{NYUv2} 
& FP      
& 32/32 
& 0.0525 & 0.9720 & 0.0459 & 0.9791 
& 0.0513 & 0.9736 & 0.0460 & 0.9770 \\
\cline{2-11}
& BRECQ    \cite{BRECQ}    
& 4/4   
& 0.2535 & 0.5395 & 0.2886 & 0.4692 
& 0.2714 & 0.5042 & 0.2910 & 0.4646 \\
& QDrop    \cite{QDROP}    
& 4/4   
& 0.1742 & 0.7166 & 0.2334 & 0.5785 
& 0.1773 & 0.7115 & 0.2332 & 0.5794 \\
& RepQ-ViT \cite{RepQ-ViT} 
& 4/4   
& 0.1959 & 0.6639 & 0.2539 & 0.5410 
& 0.2053 & 0.6385 & 0.2417 & 0.5634 \\
& QSCA     \cite{QSCA}     
& 4/4   
& \underline{0.1377} & \underline{0.8097} & \underline{0.1875} & \underline{0.6845} 
& \underline{0.1361} & \underline{0.8151} & \underline{0.1875} & \underline{0.6845} \\
& \textbf{SPA-Q (Ours)}    
& 4/4   
& \textbf{0.1090} & \textbf{0.8771} & \textbf{0.1021} & \textbf{0.8991} 
& \textbf{0.1081} & \textbf{0.8808} & \textbf{0.1001} & \textbf{0.9060} \\
\midrule
\multirow{6}{*}{KITTI}
& FP      
& 32/32 
& 0.0818 & 0.9369 & 0.0804 & 0.9396 
& 0.0832 & 0.9340 & 0.0814 & 0.9389 \\
\cline{2-11}
& BRECQ    \cite{BRECQ}    
& 4/4   
& 0.3719 & 0.3522 & 0.3989 & 0.3160 
& 0.3906 & 0.3344 & 0.3990 & 0.3175 \\
& QDrop    \cite{QDROP}    
& 4/4   
& 0.3934 & 0.3234 & 0.3855 & 0.3338 
& 0.3620 & 0.3748 & 0.3412 & 0.4082 \\
& RepQ-ViT \cite{RepQ-ViT} 
& 4/4   
& 0.3434 & 0.4159 & 0.3540 & 0.3915 
& 0.3634 & 0.3897 & 0.3906 & 0.3287 \\
& QSCA     \cite{QSCA}     
& 4/4   
& \underline{0.1874} & \underline{0.7273} & \underline{0.2365} & \underline{0.6203} 
& \underline{0.2067} & \underline{0.6794} & \underline{0.2296} & \underline{0.6174} \\ 
& \textbf{SPA-Q (Ours)} 
& 4/4 
& \textbf{0.1622} & \textbf{0.7826} & \textbf{0.1840} & \textbf{0.7230} 
& \textbf{0.1807} & \textbf{0.7449} & \textbf{0.1696} & \textbf{0.7702} \\
\midrule

\multirow{6}{*}{ETH3D}
& FP       
& 32/32 
& 0.0584 & 0.9652 & 0.0513 & 0.9741 
& 0.0548 & 0.9701 & 0.0467 & 0.9791 \\
\cline{2-11}
& BRECQ \cite{BRECQ}   
& 4/4   
& 0.2829 & 0.5082 & 0.2958 & 0.4874 
& 0.2920 & 0.4962 & 0.2963 & 0.4870 \\
& QDrop \cite{QDROP}   
& 4/4   
& 0.2351 & 0.6069 & 0.2706 & 0.5373 
& 0.2288 & 0.6186 & 0.2697 & 0.5341 \\
& RepQ-ViT \cite{RepQ-ViT}
& 4/4   
& 0.2195 & 0.6286 & 0.2491 & 0.5670 
& 0.2215 & 0.6299 & 0.2606 & 0.5538 \\
& QSCA \cite{QSCA}    
& 4/4   
& \underline{0.1791} & \underline{0.7332} & \underline{0.2241} & \underline{0.6309} 
& \underline{0.1983} & \underline{0.6791} & \underline{0.2197} & \underline{0.6298} \\
& \textbf{SPA-Q (Ours)} 
& 4/4 
& \textbf{0.1485} & \textbf{0.8071} & \textbf{0.1512} & \textbf{0.7943} 
& \textbf{0.1533} & \textbf{0.7958} & \textbf{0.1444} & \textbf{0.8140} \\
\midrule

\multirow{6}{*}{DIODE}
& FP       
& 32/32 
& 0.0753 & 0.9413 & 0.0745  & 0.9474 
& 0.0721 & 0.9426 & 0.0701 & 0.9498 \\
\cline{2-11}
& BRECQ \cite{BRECQ}   
& 4/4   
& 0.2079 & 0.6755 & 0.2201  & 0.6464 
& 0.2173 & 0.6538 & 0.2211  & 0.6441 \\
& QDrop  \cite{QDROP}  
& 4/4   
& 0.1864 & 0.7275 & 0.2058  & 0.6819 
& 0.1848 & 0.7279 & 0.2062  & 0.6777 \\
& RepQ-ViT \cite{RepQ-ViT}
& 4/4   
& 0.1746 & 0.7538 & 0.2016  & 0.6891 
& 0.1800 & 0.7552 & 0.1999  & 0.6945  \\
& QSCA \cite{QSCA}    
& 4/4   
& \underline{0.1513} & \underline{0.8033} & \underline{0.1997}  & \underline{0.7099} 
& \underline{0.1463} & \underline{0.8135} & \underline{0.1947}  & \underline{0.7094} \\
& \textbf{SPA-Q (Ours)} 
& 4/4 
& \textbf{0.1240} & \textbf{0.8653} & \textbf{0.1314} & \textbf{0.8595} 
& \textbf{0.1342} & \textbf{0.8524} & \textbf{0.1184} & \textbf{0.8901} \\
\midrule

\multirow{6}{*}{Sintel}
& FP       
& 32/32 
& 0.2296 & 0.7304 & 0.2281 & 0.7539  
& 0.2680 & 0.6974 & 0.2576 & 0.7111   \\
\cline{2-11}
& BRECQ \cite{BRECQ}   
& 4/4   
& 0.4787 & 0.3123 & 0.4820 & 0.3085  
& 0.4816 & 0.3055 & 0.4821 & 0.3008 \\
& QDrop \cite{QDROP}   
& 4/4   
& 0.4991 & 0.3181 & 0.4867 & 0.3106  
& 0.4948 & 0.3184 & 0.4882 & 0.3151 \\
& RepQ-ViT \cite{RepQ-ViT}
& 4/4   
& 0.4722 & 0.3619 & \underline{0.4642} & 0.3629  
& 0.4693 & 0.3829 & 0.4838 & 0.3237 \\
& QSCA \cite{QSCA}    
& 4/4   
& \underline{0.4531} & \underline{0.3884} & 0.4781 & \underline{0.4059}  
& \underline{0.4474} & \underline{0.3919} & \underline{0.4679} & \underline{0.4070} \\
& \textbf{SPA-Q (Ours)} 
& 4/4 
& \textbf{0.4377} & \textbf{0.4318} & \textbf{0.3949} & \textbf{0.5382} 
& \textbf{0.4211} & \textbf{0.4225} & \textbf{0.3811} & \textbf{0.4892} \\

\Xhline{1.4pt}

\end{tabular}}
\label{tab:full_table_e2e}
\end{table}

\paragraph{Implementation details.}
For fair comparison, we follow the experimental settings of prior work \cite{QSCA}. 
We adopt the widely used percentile method for calibration, using channel-wise quantization for weights and tensor-wise quantization for activations.
The calibration set is constructed by randomly sampling 16 images from the training set of NYUv2 for indoor scenes or KITTI for outdoor scenes. 
The CWDA parameters are initialized such that all elements of \(\boldsymbol{\alpha}\) are set to 1 and all elements of \(\boldsymbol{\beta}\) are set to 0. 
We optimize all CWDA parameters in an end-to-end manner for one epoch using 5\% of the training images without ground-truth depth, with \(\lambda_{\text{SILog}}=0.5\).
We use the Adam optimizer with a learning rate of \(1\times10^{-4}\) and no weight decay. All experiments are conducted with a batch size of 1 on a single RTX 6000 Ada Generation GPU.

\subsection{Experimental Results}
\paragraph{Quantitative results.}
As shown in Table~\ref{tab:full_table_e2e}, we compare SPA-Q with existing PTQ methods, including BRECQ \cite{BRECQ}, QDrop \cite{QDROP}, RepQ-ViT \cite{RepQ-ViT}, and QSCA \cite{QSCA}, under 4-bit quantization on five zero-shot relative depth estimation benchmarks. 
Overall, our framework consistently achieves the best performance in both AbsRel and $\delta_1$ across all evaluated datasets, outperforming all compared PTQ methods. For Depth Anything v1 with the ViT-B backbone, the improvement is substantial on the two representative benchmarks, NYUv2 \cite{nyu} and KITTI \cite{kitti}. 
On NYUv2, our framework reduces AbsRel from 0.1875 to 0.1021 and improves $\delta_1$ from 0.6845 to 0.8991. On KITTI, AbsRel is further reduced from 0.2365 to 0.1840 and $\delta_1$ is improved from 0.6203 to 0.7230. 
\begin{figure}[t]
    \centering
    \includegraphics[width=0.95\textwidth]{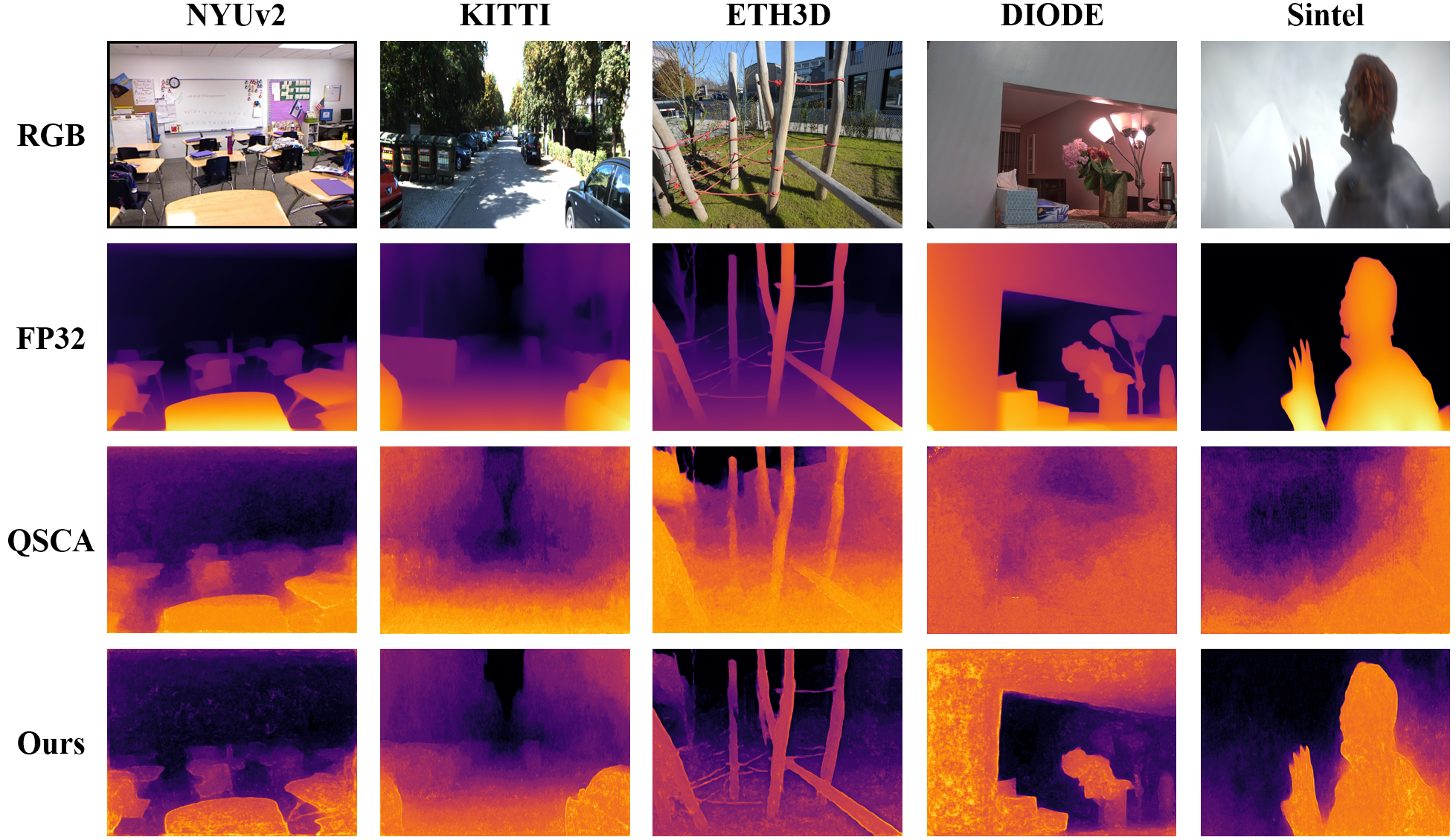}
    \caption{Qualitative comparison of 4-bit quantization results for Depth Anything with a ViT-S backbone on benchmarks: NYUv2 \cite{nyu}, KITTI \cite{kitti}, ETH3D \cite{eth3d}, DIODE \cite{diode}, and Sintel \cite{sintel}.}
    \label{fig:fig_vis} 
\end{figure}
These results show that the proposed framework effectively preserves depth estimation performance under severe low-bit quantization in both indoor and outdoor settings. 
Consistent improvements are also observed on ETH3D \cite{eth3d}, DIODE \cite{diode}, and Sintel \cite{sintel}, where SPA-Q achieves the best results among the compared PTQ methods.
In particular, the improvement remains consistent on Sintel, a challenging synthetic benchmark, indicating that the proposed framework generalizes well not only to indoor and outdoor scenes but also to synthetic domains under 4-bit quantization.

\paragraph{Qualitative results.}
To validate the effectiveness of the proposed framework, we present qualitative comparisons of predicted depth maps in Figure~\ref{fig:fig_vis}. 
Compared with QSCA \cite{QSCA}, SPA-Q produces qualitatively better depth maps with improved structural consistency and sharper object boundaries across diverse indoor, outdoor, and synthetic scenes. 
For NYUv2 \cite{nyu}, it better captures the indoor layout, including the desk and chair boundaries, while providing clearer separation between foreground objects and distant regions. 
On KITTI \cite{kitti}, it produces smoother depth transitions across the road scene, with better separation between nearby vehicles and the background. 
Similarly, on ETH3D \cite{eth3d} and DIODE \cite{diode}, thin structures such as wooden poles and ropes, as well as the boundary of the wall opening and object silhouettes, are well recovered, whereas QSCA largely fails to preserve them.
Sintel \cite{sintel} is particularly challenging due to its domain gap from real-world imagery and rendering effects such as atmospheric haze and motion blur.
While QSCA collapses into a nearly radial depth pattern that loses the character entirely, SPA-Q recovers its contours, including the raised hand, consistent with the full-precision prediction.
These results demonstrate that the proposed framework consistently preserves depth structure across diverse domains, including indoor, outdoor, and synthetic scenes.
\begin{figure}[t]
    \centering
    \includegraphics[width=0.9\textwidth]{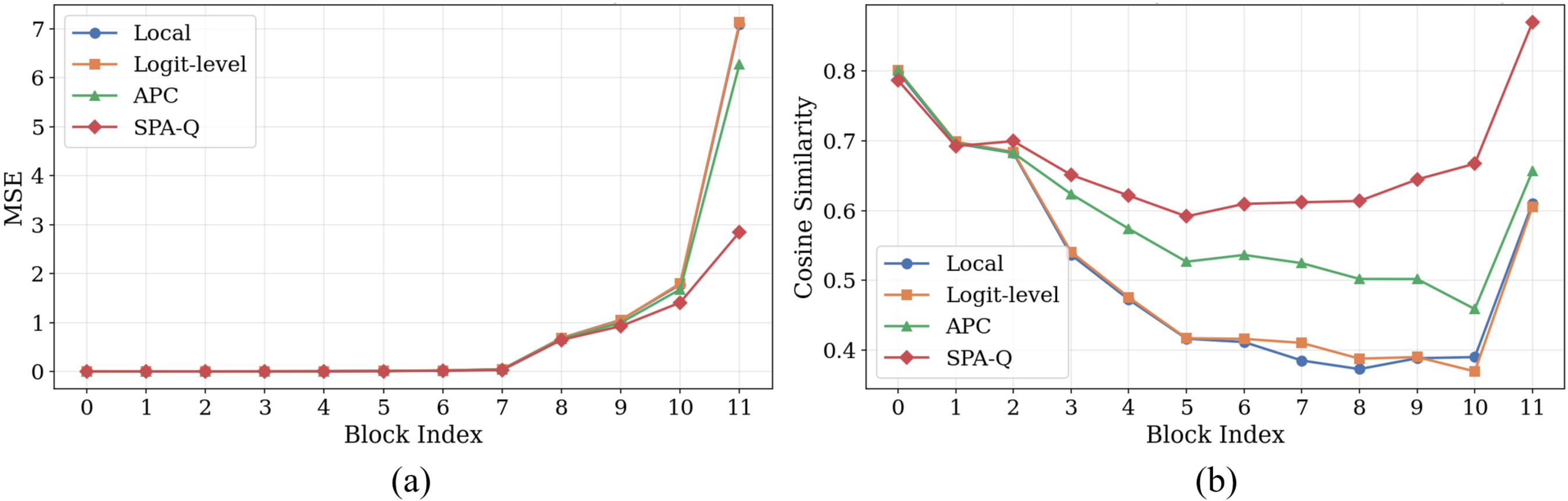}
    \caption{Block-wise feature analysis under 4-bit quantization for Depth Anything v1 \cite{DAv1} with a ViT-B backbone across transformer blocks. (a) Mean squared error (MSE) between full-precision and quantized features. (b) Cosine similarity between full-precision and quantized features.}
    \label{fig:fig_ablation}
\end{figure}
\begin{table}[t]
\centering
\begin{minipage}[t]{0.43\textwidth}
\centering
\caption{Ablation study of calibration objectives under 4-bit quantization for Depth Anything v1 with a ViT-B backbone on NYUv2 \cite{nyu} and KITTI \cite{kitti}.}
\label{tab:ablation_1}
\end{minipage}
\hfill
\begin{minipage}[t]{0.54\textwidth}
\centering
\caption{Comparison of model size, parameter count, and bit operations (BOPs) under 4-bit quantization. SPA-Q* denotes the model before fusing the learned CWDA parameters into the weights.}
\label{tab:efficiency}
\end{minipage}

\begin{minipage}[t]{0.43\textwidth}
\centering
\footnotesize
\renewcommand{\arraystretch}{1.0}
\begin{tabular*}{\linewidth}{@{\extracolsep{\fill}} c l c c @{}}
\toprule
\textbf{Dataset} & \textbf{Method} & \textbf{AbsRel $\downarrow$} & \textbf{$\delta_1$ $\uparrow$} \\
\midrule
\multirow{4}{*}{NYUv2}
& Local       & 0.1908 & 0.6836 \\
& Logits-level & 0.1839 & 0.7013 \\
& APC         & 0.1444 & 0.7990 \\
& SPA-Q       & \textbf{0.1021} & \textbf{0.8991} \\
\midrule
\multirow{4}{*}{KITTI}
& Local       & 0.3241 & 0.4366 \\
& Logits-level & 0.3002 & 0.4828 \\
& APC         & 0.2490 & 0.5789 \\
& SPA-Q       & \textbf{0.1840} & \textbf{0.7230} \\
\bottomrule
\end{tabular*}
\end{minipage}
\hfill
\begin{minipage}[t]{0.54\textwidth}
\centering
\footnotesize
\setlength{\tabcolsep}{2.5pt}
\renewcommand{\arraystretch}{1.0}
\begin{tabular*}{\linewidth}{@{\extracolsep{\fill}} c l c c c @{}}
\toprule
\textbf{Encoder} & \textbf{Method} & \textbf{Model size (MB)} & \textbf{Params (M)} & \textbf{BOPs (G)} \\
\midrule
\multirow{4}{*}{ViT-S}
& FP                & 94.55 & 24.79 & 85742 \\
& QSCA \cite{QSCA}  & 16.27 & 25.23 & 1544  \\
& SPA-Q*            & 14.50 & 24.87 & 1457  \\
& SPA-Q             & 14.16 & 24.79 & 1340  \\
\midrule
\multirow{4}{*}{ViT-B}
& FP                & 371.82 & 97.47 & 273280 \\
& QSCA \cite{QSCA}  & 60.02  & 99.24 & 5092   \\
& SPA-Q*            & 51.84  & 97.65 & 4493   \\
& SPA-Q             & 51.16  & 97.47 & 4270   \\
\bottomrule
\end{tabular*}
\end{minipage}
\end{table}

\subsection{Ablation Study}
\paragraph{Comparison of calibration objectives.}
We conduct an ablation study on NYUv2 \cite{nyu} and KITTI \cite{kitti} to examine the effect of the proposed calibration objective under 4-bit quantization for Depth Anything v1 with a ViT-B backbone. 
As shown in Table~\ref{tab:ablation_1}, on NYUv2, replacing local quantization error minimization with a logits-level objective yields only a marginal improvement, with AbsRel changing from 0.1908 to 0.1839 and $\delta_1$ from 0.6836 to 0.7013. 
This result suggests that directly considering query-key interaction at the logits level is still insufficient to consistently improve final depth performance. 
In contrast, replacing the logits-level objective with APC further improves performance, reducing AbsRel to 0.1444 and improving $\delta_1$ to 0.7990.
Finally, combining APC with CWDA yields the best performance, reducing AbsRel to 0.1021 and improving $\delta_1$ to 0.8991. 
KITTI shows a similar trend, where logits-level calibration provides only limited gains, while APC and the full SPA-Q framework lead to clear performance improvements.

\paragraph{Block-wise comparison of feature similarity.}
We further evaluate whether SPA-Q preserves encoder features for dense depth prediction across transformer blocks.
Since the decoder reconstructs depth maps from encoder features, preserving their fidelity is important for maintaining depth structure under quantization.
As shown in Figure~\ref{fig:fig_ablation} (a), SPA-Q consistently achieves the lowest MSE across most blocks, with the advantage becoming more pronounced in deeper blocks.
Figure~\ref{fig:fig_ablation} (b) shows that SPA-Q yields the highest cosine similarity to the full-precision features throughout the encoder, indicating better block-wise feature preservation.
These results demonstrate that APC and CWDA jointly preserve the encoder features required for depth prediction by improving attention-aware calibration and mitigating accumulated feature distortion.

\paragraph{Efficiency analysis.}
We further compare model size, parameter count, and bit operations (BOPs) of SPA-Q and QSCA \cite{QSCA} under 4-bit quantization for Depth Anything v1 with ViT-S and ViT-B backbones. The efficiency results in Table~\ref{tab:efficiency} are reported on the NYUv2 dataset. As shown in Table~\ref{tab:efficiency}, SPA-Q is more efficient than QSCA in both model size and computational cost. For ViT-S, the fused SPA-Q model reduces model size from 16.27 MB to 14.16 MB and BOPs from 1544 G to 1340 G. For ViT-B, SPA-Q also yields a smaller model size than QSCA, decreasing it from 60.02 MB to 51.16 MB after fusion. Although SPA-Q before fusion introduces additional channel-wise affine parameters, these parameters are absorbed into the weights after training, so the fused model restores the parameter count to that of the original quantized network. These results show that SPA-Q improves depth estimation accuracy without introducing additional inference overhead.

\section{Conclusion}

In this paper, we addressed low-bit post-training quantization for MDE.
We identified two key limitations of existing PTQ methods.
Independently minimizing quantization error for query and key projections does not adequately preserve the resulting attention distribution, while accumulated quantization errors induce distribution shifts in intermediate features.
To address these issues, we proposed SPA-Q, a structure-preserving adaptive PTQ framework that combines APC and CWDA to preserve attention distributions and mitigate feature distortion.
Experimental results on multiple zero-shot depth estimation benchmarks showed that SPA-Q consistently outperforms existing PTQ methods under 4-bit quantization.

\paragraph{Limitations \& Future Works.}
Although the proposed method consistently outperforms existing PTQ baselines under 4-bit quantization and yields improved qualitative results, the quantized depth maps remain visibly less clean than their full-precision counterparts. 
We conjecture that this limitation stems from the restricted set of depth values representable under low-bit quantization, which is more discrete than in full precision and can reduce the smoothness of predicted depth maps.
Our study is limited to MDE with Depth Anything v1 and v2, partly due to GPU memory constraints for larger foundation models.
Extending SPA-Q to larger foundation models and other depth estimation settings, such as stereo and video depth estimation, remains an important direction for future work.

\section*{Acknowledgements}

This work was supported by the Institute of Information \& Communications Technology Planning \& Evaluation(IITP) grant funded by the Korea government(MSIT) (No.RS-2025-02263706, Development of an analog-digital mixed ultra-low power neuromorphic edge SoC, 25\%) and the IITP(Institute of Information \& Communications Technology Planning \& Evaluation)-ITRC(Information Technology Research Center) grant funded by the Korea government(Ministry of Science and ICT) (IITP-2026-RS-2023-00260091, 25\%) and Institute of Information communications Technology Planning Evaluation (IITP) under the Artificial Intelligence Innovation Human Resources Development (IITP-RS-2026-25547954, 25\%) grant funded by the Korea government(MSIT) and the K-On-Device AI Semiconductor RD Program (RS-2026-25620710, Development of On-Device AI Semiconductor-Based AI Algorithms, Intelligence Interconnection Framework, and Software Platforms for AI Operations, 25\%) funded By the Ministry of Trade, Industry and Energy(MOTIE, Korea).

{
    \small
    \bibliographystyle{unsrtnat}
    \bibliography{citation}

\begin{thebibliography}{47}
\providecommand{\natexlab}[1]{#1}
\providecommand{\url}[1]{\texttt{#1}}
\expandafter\ifx\csname urlstyle\endcsname\relax
  \providecommand{\doi}[1]{doi: #1}\else
  \providecommand{\doi}{doi: \begingroup \urlstyle{rm}\Url}\fi

\bibitem[Dong et~al.(2022)Dong, Garratt, Anavatti, and Abbass]{survey}
Xingshuai Dong, Matthew~A Garratt, Sreenatha~G Anavatti, and Hussein~A Abbass.
\newblock Towards real-time monocular depth estimation for robotics: A survey.
\newblock \emph{IEEE Transactions on Intelligent Transportation Systems},
  23\penalty0 (10):\penalty0 16940--16961, 2022.

\bibitem[Wofk et~al.(2019)Wofk, Ma, Yang, Karaman, and Sze]{fastdepth}
Diana Wofk, Fangchang Ma, Tien-Ju Yang, Sertac Karaman, and Vivienne Sze.
\newblock Fastdepth: Fast monocular depth estimation on embedded systems.
\newblock In \emph{2019 International Conference on Robotics and Automation
  (ICRA)}, pages 6101--6108. IEEE, 2019.

\bibitem[Weng and Kitani(2019)]{autonomous1}
Xinshuo Weng and Kris Kitani.
\newblock Monocular 3d object detection with pseudo-lidar point cloud.
\newblock In \emph{Proceedings of the IEEE/CVF international conference on
  computer vision workshops}, 2019.

\bibitem[Wang et~al.(2019)Wang, Chao, Garg, Hariharan, Campbell, and
  Weinberger]{autonomous2}
Yan Wang, Wei-Lun Chao, Divyansh Garg, Bharath Hariharan, Mark Campbell, and
  Kilian~Q Weinberger.
\newblock Pseudo-lidar from visual depth estimation: Bridging the gap in 3d
  object detection for autonomous driving.
\newblock In \emph{Proceedings of the IEEE/CVF conference on computer vision
  and pattern recognition}, pages 8445--8453, 2019.

\bibitem[Mildenhall et~al.(2021)Mildenhall, Srinivasan, Tancik, Barron,
  Ramamoorthi, and Ng]{nerf}
Ben Mildenhall, Pratul~P Srinivasan, Matthew Tancik, Jonathan~T Barron, Ravi
  Ramamoorthi, and Ren Ng.
\newblock Nerf: Representing scenes as neural radiance fields for view
  synthesis.
\newblock \emph{Communications of the ACM}, 65\penalty0 (1):\penalty0 99--106,
  2021.

\bibitem[Kerbl et~al.(2023)Kerbl, Kopanas, Leimk{\"u}hler, Drettakis,
  et~al.]{gaussian}
Bernhard Kerbl, Georgios Kopanas, Thomas Leimk{\"u}hler, George Drettakis,
  et~al.
\newblock 3d gaussian splatting for real-time radiance field rendering.
\newblock \emph{ACM Trans. Graph.}, 42\penalty0 (4):\penalty0 139:1--139:14,
  2023.

\bibitem[Yang et~al.(2025)Yang, Zinke, Kang, Cho, Choi, Lee, and Kang]{TII}
Jincheol Yang, Matti Zinke, Beoungwoo Kang, Hyung~Uk Cho, Hyunyoung Choi,
  SangGu Lee, and Suk-Ju Kang.
\newblock Efficient monocular depth-based physical distance measurement for
  low-depth scales.
\newblock \emph{IEEE Transactions on Industrial Informatics}, 2025.

\bibitem[Dosovitskiy et~al.(2020)Dosovitskiy, Beyer, Kolesnikov, Weissenborn,
  Zhai, Unterthiner, Dehghani, Minderer, Heigold, Gelly, et~al.]{vit}
Alexey Dosovitskiy, Lucas Beyer, Alexander Kolesnikov, Dirk Weissenborn,
  Xiaohua Zhai, Thomas Unterthiner, Mostafa Dehghani, Matthias Minderer, Georg
  Heigold, Sylvain Gelly, et~al.
\newblock An image is worth 16x16 words: Transformers for image recognition at
  scale.
\newblock \emph{arXiv preprint arXiv:2010.11929}, 2020.

\bibitem[Touvron et~al.(2021)Touvron, Cord, Douze, Massa, Sablayrolles, and
  J{\'e}gou]{deit}
Hugo Touvron, Matthieu Cord, Matthijs Douze, Francisco Massa, Alexandre
  Sablayrolles, and Herv{\'e} J{\'e}gou.
\newblock Training data-efficient image transformers \& distillation through
  attention.
\newblock In \emph{International conference on machine learning}, pages
  10347--10357. PMLR, 2021.

\bibitem[Caron et~al.(2021)Caron, Touvron, Misra, J{\'e}gou, Mairal,
  Bojanowski, and Joulin]{DINOv1}
Mathilde Caron, Hugo Touvron, Ishan Misra, Herv{\'e} J{\'e}gou, Julien Mairal,
  Piotr Bojanowski, and Armand Joulin.
\newblock Emerging properties in self-supervised vision transformers.
\newblock In \emph{Proceedings of the IEEE/CVF international conference on
  computer vision}, pages 9650--9660, 2021.

\bibitem[Oquab et~al.(2023)Oquab, Darcet, Moutakanni, Vo, Szafraniec, Khalidov,
  Fernandez, Haziza, Massa, El-Nouby, et~al.]{DINOv2}
Maxime Oquab, Timoth{\'e}e Darcet, Th{\'e}o Moutakanni, Huy Vo, Marc
  Szafraniec, Vasil Khalidov, Pierre Fernandez, Daniel Haziza, Francisco Massa,
  Alaaeldin El-Nouby, et~al.
\newblock Dinov2: Learning robust visual features without supervision.
\newblock \emph{arXiv preprint arXiv:2304.07193}, 2023.

\bibitem[Yu et~al.(2024)Yu, Cho, Kang, Moon, Kong, and Kang]{embedding}
Hyunwoo Yu, Yubin Cho, Beoungwoo Kang, Seunghun Moon, Kyeongbo Kong, and Suk-Ju
  Kang.
\newblock Embedding-free transformer with inference spatial reduction for
  efficient semantic segmentation.
\newblock In \emph{European Conference on Computer Vision}, pages 92--110.
  Springer, 2024.

\bibitem[Ranftl et~al.(2022)Ranftl, Lasinger, Hafner, Schindler, and
  Koltun]{MiDaS}
Ren{\'e} Ranftl, Katrin Lasinger, David Hafner, Konrad Schindler, and Vladlen
  Koltun.
\newblock Towards robust monocular depth estimation: Mixing datasets for
  zero-shot cross-dataset transfer.
\newblock \emph{IEEE transactions on pattern analysis and machine
  intelligence}, 44\penalty0 (3):\penalty0 1623--1637, 2022.

\bibitem[Yang et~al.(2024{\natexlab{a}})Yang, Kang, Huang, Xu, Feng, and
  Zhao]{DAv1}
Lihe Yang, Bingyi Kang, Zilong Huang, Xiaogang Xu, Jiashi Feng, and Hengshuang
  Zhao.
\newblock Depth anything: Unleashing the power of large-scale unlabeled data.
\newblock In \emph{Proceedings of the IEEE/CVF conference on computer vision
  and pattern recognition}, pages 10371--10381, 2024{\natexlab{a}}.

\bibitem[Yang et~al.(2024{\natexlab{b}})Yang, Kang, Huang, Zhao, Xu, Feng, and
  Zhao]{DAv2}
Lihe Yang, Bingyi Kang, Zilong Huang, Zhen Zhao, Xiaogang Xu, Jiashi Feng, and
  Hengshuang Zhao.
\newblock Depth anything v2.
\newblock \emph{Advances in Neural Information Processing Systems},
  37:\penalty0 21875--21911, 2024{\natexlab{b}}.

\bibitem[Gholami et~al.(2022)Gholami, Kim, Dong, Yao, Mahoney, and
  Keutzer]{quant_survey}
Amir Gholami, Sehoon Kim, Zhen Dong, Zhewei Yao, Michael~W Mahoney, and Kurt
  Keutzer.
\newblock A survey of quantization methods for efficient neural network
  inference.
\newblock In \emph{Low-power computer vision}, pages 291--326. Chapman and
  Hall/CRC, 2022.

\bibitem[Nagel et~al.(2021)Nagel, Fournarakis, Amjad, Bondarenko, Van~Baalen,
  and Blankevoort]{white}
Markus Nagel, Marios Fournarakis, Rana~Ali Amjad, Yelysei Bondarenko, Mart
  Van~Baalen, and Tijmen Blankevoort.
\newblock A white paper on neural network quantization.
\newblock \emph{arXiv preprint arXiv:2106.08295}, 2021.

\bibitem[Wu et~al.(2020)Wu, Judd, Zhang, Isaev, and Micikevicius]{integer}
Hao Wu, Patrick Judd, Xiaojie Zhang, Mikhail Isaev, and Paulius Micikevicius.
\newblock Integer quantization for deep learning inference: Principles and
  empirical evaluation.
\newblock \emph{arXiv preprint arXiv:2004.09602}, 2020.

\bibitem[Choukroun et~al.(2019)Choukroun, Kravchik, Yang, and Kisilev]{low}
Yoni Choukroun, Eli Kravchik, Fan Yang, and Pavel Kisilev.
\newblock Low-bit quantization of neural networks for efficient inference.
\newblock In \emph{2019 IEEE/CVF International Conference on Computer Vision
  Workshop (ICCVW)}, pages 3009--3018. IEEE, 2019.

\bibitem[Nagel et~al.(2020)Nagel, Amjad, Van~Baalen, Louizos, and
  Blankevoort]{adaround}
Markus Nagel, Rana~Ali Amjad, Mart Van~Baalen, Christos Louizos, and Tijmen
  Blankevoort.
\newblock Up or down? adaptive rounding for post-training quantization.
\newblock In \emph{International conference on machine learning}, pages
  7197--7206. PMLR, 2020.

\bibitem[Lin et~al.(2021)Lin, Zhang, Sun, Li, and Zhou]{FQ-ViT}
Yang Lin, Tianyu Zhang, Peiqin Sun, Zheng Li, and Shuchang Zhou.
\newblock Fq-vit: Post-training quantization for fully quantized vision
  transformer.
\newblock \emph{arXiv preprint arXiv:2111.13824}, 2021.

\bibitem[Esser et~al.(2019)Esser, McKinstry, Bablani, Appuswamy, and
  Modha]{LSQ}
Steven~K Esser, Jeffrey~L McKinstry, Deepika Bablani, Rathinakumar Appuswamy,
  and Dharmendra~S Modha.
\newblock Learned step size quantization.
\newblock \emph{arXiv preprint arXiv:1902.08153}, 2019.

\bibitem[Bhalgat et~al.(2020)Bhalgat, Lee, Nagel, Blankevoort, and Kwak]{LSQ+}
Yash Bhalgat, Jinwon Lee, Markus Nagel, Tijmen Blankevoort, and Nojun Kwak.
\newblock Lsq+: Improving low-bit quantization through learnable offsets and
  better initialization.
\newblock In \emph{Proceedings of the IEEE/CVF conference on computer vision
  and pattern recognition workshops}, pages 696--697, 2020.

\bibitem[Choi et~al.(2018)Choi, Wang, Venkataramani, Chuang, Srinivasan, and
  Gopalakrishnan]{PACT}
Jungwook Choi, Zhuo Wang, Swagath Venkataramani, Pierce I-Jen Chuang,
  Vijayalakshmi Srinivasan, and Kailash Gopalakrishnan.
\newblock Pact: Parameterized clipping activation for quantized neural
  networks.
\newblock \emph{arXiv preprint arXiv:1805.06085}, 2018.

\bibitem[Li et~al.(2021)Li, Gong, Tan, Yang, Hu, Zhang, Yu, Wang, and
  Gu]{BRECQ}
Yuhang Li, Ruihao Gong, Xu~Tan, Yang Yang, Peng Hu, Qi~Zhang, Fengwei Yu, Wei
  Wang, and Shi Gu.
\newblock Brecq: Pushing the limit of post-training quantization by block
  reconstruction.
\newblock \emph{arXiv preprint arXiv:2102.05426}, 2021.

\bibitem[Wei et~al.(2022)Wei, Gong, Li, Liu, and Yu]{QDROP}
Xiuying Wei, Ruihao Gong, Yuhang Li, Xianglong Liu, and Fengwei Yu.
\newblock Qdrop: Randomly dropping quantization for extremely low-bit
  post-training quantization.
\newblock \emph{arXiv preprint arXiv:2203.05740}, 2022.

\bibitem[Li et~al.(2023)Li, Xiao, Yang, and Gu]{RepQ-ViT}
Zhikai Li, Junrui Xiao, Lianwei Yang, and Qingyi Gu.
\newblock Repq-vit: Scale reparameterization for post-training quantization of
  vision transformers.
\newblock In \emph{Proceedings of the IEEE/CVF International Conference on
  Computer Vision}, pages 17227--17236, 2023.

\bibitem[Zhong et~al.(2024)Zhong, Hu, Huang, Zhang, and Ji]{erq}
Yunshan Zhong, Jiawei Hu, You Huang, Yuxin Zhang, and Rongrong Ji.
\newblock Erq: Error reduction for post-training quantization of vision
  transformers.
\newblock In \emph{Forty-First International Conference on Machine Learning},
  2024.

\bibitem[Lee et~al.(2019)Lee, Han, Ko, and Suh]{bts}
Jin~Han Lee, Myung-Kyu Han, Dong~Wook Ko, and Il~Hong Suh.
\newblock From big to small: Multi-scale local planar guidance for monocular
  depth estimation.
\newblock \emph{arXiv preprint arXiv:1907.10326}, 2019.

\bibitem[Bhat et~al.(2023)Bhat, Birkl, Wofk, Wonka, and M{\"u}ller]{zoedepth}
Shariq~Farooq Bhat, Reiner Birkl, Diana Wofk, Peter Wonka, and Matthias
  M{\"u}ller.
\newblock Zoedepth: Zero-shot transfer by combining relative and metric depth.
\newblock \emph{arXiv preprint arXiv:2302.12288}, 2023.

\bibitem[Ranftl et~al.(2021)Ranftl, Bochkovskiy, and Koltun]{DPT}
Ren{\'e} Ranftl, Alexey Bochkovskiy, and Vladlen Koltun.
\newblock Vision transformers for dense prediction.
\newblock In \emph{Proceedings of the IEEE/CVF international conference on
  computer vision}, pages 12179--12188, 2021.

\bibitem[Liu et~al.(2021)Liu, Lin, Cao, Hu, Wei, Zhang, Lin, and Guo]{swin}
Ze~Liu, Yutong Lin, Yue Cao, Han Hu, Yixuan Wei, Zheng Zhang, Stephen Lin, and
  Baining Guo.
\newblock Swin transformer: Hierarchical vision transformer using shifted
  windows.
\newblock In \emph{Proceedings of the IEEE/CVF international conference on
  computer vision}, pages 10012--10022, 2021.

\bibitem[Yang et~al.(2026)Yang, Choi, Zinke, and Kang]{QSCA}
Jincheol Yang, Jaemin Choi, Matti Zinke, and Suk-Ju Kang.
\newblock Qsca: Quantization with self-compensating auxiliary for monocular
  depth estimation.
\newblock \emph{Advances in Neural Information Processing Systems},
  38:\penalty0 31444--31465, 2026.

\bibitem[Yang et~al.(2024{\natexlab{c}})Yang, Gong, Lin, Wu, Sun, and Gu]{dopq}
Lianwei Yang, Haisong Gong, Haokun Lin, Yichen Wu, Zhenan Sun, and Qingyi Gu.
\newblock Dopq-vit: Towards distribution-friendly and outlier-aware
  post-training quantization for vision transformers.
\newblock \emph{arXiv preprint arXiv:2408.03291}, 2024{\natexlab{c}}.

\bibitem[Jiang et~al.(2026)Jiang, Jiang, Yao, Cheng, and Han]{uq-vit}
Tao Jiang, Yucheng Jiang, Xiwen Yao, Gong Cheng, and Junwei Han.
\newblock Uq-vit: Harmonizing extreme activations with hardware-friendly
  uniform quantization in vision transformers.
\newblock In \emph{Proceedings of the AAAI Conference on Artificial
  Intelligence}, volume~40, pages 22354--22362, 2026.

\bibitem[Silberman et~al.(2012)Silberman, Hoiem, Kohli, and Fergus]{nyu}
Nathan Silberman, Derek Hoiem, Pushmeet Kohli, and Rob Fergus.
\newblock Indoor segmentation and support inference from rgbd images.
\newblock In \emph{European conference on computer vision}, pages 746--760.
  Springer, 2012.

\bibitem[Geiger et~al.(2013)Geiger, Lenz, Stiller, and Urtasun]{kitti}
Andreas Geiger, Philip Lenz, Christoph Stiller, and Raquel Urtasun.
\newblock Vision meets robotics: The kitti dataset.
\newblock \emph{The international journal of robotics research}, 32\penalty0
  (11):\penalty0 1231--1237, 2013.

\bibitem[Schops et~al.(2017)Schops, Schonberger, Galliani, Sattler, Schindler,
  Pollefeys, and Geiger]{eth3d}
Thomas Schops, Johannes~L Schonberger, Silvano Galliani, Torsten Sattler,
  Konrad Schindler, Marc Pollefeys, and Andreas Geiger.
\newblock A multi-view stereo benchmark with high-resolution images and
  multi-camera videos.
\newblock In \emph{Proceedings of the IEEE conference on computer vision and
  pattern recognition}, pages 3260--3269, 2017.

\bibitem[Vasiljevic et~al.(2019)Vasiljevic, Kolkin, Zhang, Luo, Wang, Dai,
  Daniele, Mostajabi, Basart, Walter, et~al.]{diode}
Igor Vasiljevic, Nick Kolkin, Shanyi Zhang, Ruotian Luo, Haochen Wang, Falcon~Z
  Dai, Andrea~F Daniele, Mohammadreza Mostajabi, Steven Basart, Matthew~R
  Walter, et~al.
\newblock Diode: A dense indoor and outdoor depth dataset.
\newblock \emph{arXiv preprint arXiv:1908.00463}, 2019.

\bibitem[Butler et~al.(2012)Butler, Wulff, Stanley, and Black]{sintel}
Daniel~J Butler, Jonas Wulff, Garrett~B Stanley, and Michael~J Black.
\newblock A naturalistic open source movie for optical flow evaluation.
\newblock In \emph{European conference on computer vision}, pages 611--625.
  Springer, 2012.

\bibitem[Li et~al.(2019)Li, Wang, Liang, Qin, Yan, and Fan]{Percentile}
Rundong Li, Yan Wang, Feng Liang, Hongwei Qin, Junjie Yan, and Rui Fan.
\newblock Fully quantized network for object detection.
\newblock In \emph{Proceedings of the IEEE/CVF conference on computer vision
  and pattern recognition}, pages 2810--2819, 2019.

\bibitem[Yuan et~al.(2022)Yuan, Xue, Chen, Wu, and Sun]{ptq4vit}
Zhihang Yuan, Chenhao Xue, Yiqi Chen, Qiang Wu, and Guangyu Sun.
\newblock Ptq4vit: Post-training quantization for vision transformers with twin
  uniform quantization.
\newblock In \emph{European conference on computer vision}, pages 191--207.
  Springer, 2022.

\bibitem[Fu et~al.(2025)Fu, Yu, Shao, Zhou, Zhu, and Wu]{qwt}
Minghao Fu, Hao Yu, Jie Shao, Junjie Zhou, Ke~Zhu, and Jianxin Wu.
\newblock Quantization without tears.
\newblock In \emph{Proceedings of the Computer Vision and Pattern Recognition
  Conference}, pages 4462--4472, 2025.

\bibitem[Hu et~al.(2019)Hu, Ozay, Zhang, and Okatani]{EdgeF1}
Junjie Hu, Mete Ozay, Yan Zhang, and Takayuki Okatani.
\newblock Revisiting single image depth estimation: Toward higher resolution
  maps with accurate object boundaries.
\newblock In \emph{2019 IEEE winter conference on applications of computer
  vision (WACV)}, pages 1043--1051. IEEE, 2019.

\bibitem[Bochkovskiy et~al.(2025)Bochkovskiy, Delaunoy, Germain, Santos, Zhou,
  Richter, and Koltun]{BoundaryF1}
Alexey Bochkovskiy, Ama{\"e}l Delaunoy, Hugo Germain, Marcel Santos, Yichao
  Zhou, Stephan Richter, and Vladlen Koltun.
\newblock Depth pro: Sharp monocular metric depth in less than a second.
\newblock In \emph{International Conference on Learning Representations},
  volume 2025, pages 75602--75637, 2025.

\bibitem[Xian et~al.(2020)Xian, Zhang, Wang, Mai, Lin, and Cao]{ORD}
Ke~Xian, Jianming Zhang, Oliver Wang, Long Mai, Zhe Lin, and Zhiguo Cao.
\newblock Structure-guided ranking loss for single image depth prediction.
\newblock In \emph{2020 IEEE/CVF Conference on Computer Vision and Pattern
  Recognition (CVPR)}, pages 608--617. IEEE, 2020.

\bibitem[Koch et~al.(2018)Koch, Liebel, Fraundorfer, and K{\"o}rner]{iBims-1}
Tobias Koch, Lukas Liebel, Friedrich Fraundorfer, and Marco K{\"o}rner.
\newblock Evaluation of cnn-based single-image depth estimation methods.
\newblock In \emph{European Conference on Computer Vision}, pages 331--348.
  Springer, 2018.

\end{thebibliography}
}

\appendix
\newpage
\section*{Appendix}

\section{Extended Comparison with Additional PTQ Baselines}

\begin{table}[H]
\centering
\caption{Extended comparison with additional PTQ baselines on NYUv2 \cite{nyu} and KITTI \cite{kitti} for zero-shot relative depth estimation under 4-bit quantization.
\textit{E.} denotes the encoder backbone used in the MDE architectures.}

\setlength{\tabcolsep}{10pt}
\renewcommand{\arraystretch}{1.5}
\resizebox{\textwidth}{!}{
\begin{tabular}{c l c c c c c c c c c}
\Xhline{1.4pt}
\multirow{3}{*}{\textbf{Dataset}} & \multirow{3}{*}{\textbf{Method}} & \multirow{3}{*}{\textbf{W/A}}
& \multicolumn{4}{c}{\textbf{Depth Anything v1}} 
& \multicolumn{4}{c}{\textbf{Depth Anything v2}} \\
\cmidrule(lr){4-7} \cmidrule(lr){8-11}
& & 
& \multicolumn{2}{c}{\textbf{\textit{E.} ViT-S}} 
& \multicolumn{2}{c}{\textbf{\textit{E.} ViT-B}} 
& \multicolumn{2}{c}{\textbf{\textit{E.} ViT-S}} 
& \multicolumn{2}{c}{\textbf{\textit{E.} ViT-B}} \\
& & 
& AbsRel $\downarrow$ & $\delta_1 \uparrow$
& AbsRel $\downarrow$ & $\delta_1 \uparrow$
& AbsRel $\downarrow$ & $\delta_1 \uparrow$
& AbsRel $\downarrow$ & $\delta_1 \uparrow$ \\
\midrule
\multirow{11}{*}{NYUv2} 
& FP      & 32/32 & 0.0525 & 0.9720 & 0.0459 & 0.9791 & 0.0513 & 0.9736 & 0.0460 & 0.9770 \\
\cline{2-11}
& MinMax \cite{quant_survey} & 4/4 & 0.2728 & 0.5024 & 1.5735 & 0.1972 & 0.2815 & 0.4873 & 2.3712 & 0.1102 \\
& Percentile \cite{Percentile} & 4/4 & 0.2006 & 0.6542 & 0.2522 & 0.5430 & 0.1935 & 0.6675 & 0.2543 & 0.5379\\
& BRECQ    \cite{BRECQ}    & 4/4   & 0.2535 & 0.5395 & 0.2886 & 0.4692 & 0.2714 & 0.5042 & 0.2910 & 0.4646 \\
& QDrop    \cite{QDROP}    & 4/4   & 0.1742 & 0.7166 & 0.2334 & 0.5785 & 0.1773 & 0.7115 & 0.2332 & 0.5794 \\
& PTQ4ViT \cite{ptq4vit} & 4/4 & 0.2393 & 0.5693 & 0.2294 & 0.5895 & 0.2735 & 0.5034 & 0.2694 & 0.5183 \\
& RepQ-ViT \cite{RepQ-ViT} & 4/4   & 0.1959 & 0.6639 & 0.2539 & 0.5410 & 0.2053 & 0.6385 & 0.2417 & 0.5634 \\
& ERQ \cite{erq} & 4/4 & 0.1785 & 0.7062 & 0.2182 & 0.6126 & 0.1751 & 0.7140 & 0.2883 & 0.4705\\
& QwT \cite{qwt} & 4/4 & 0.1407 & 0.8007 & 0.2024 & 0.6486 & 0.1400 & 0.8050 & 0.1992 & 0.6589\\
& QSCA     \cite{QSCA}     & 4/4   & \underline{0.1377} & \underline{0.8097} & \underline{0.1875} & \underline{0.6845} & \underline{0.1361} & \underline{0.8151} & \underline{0.1875} & \underline{0.6845} \\
& \textbf{SPA-Q (Ours)}    & 4/4   & \textbf{0.1090} & \textbf{0.8771} & \textbf{0.1021} & \textbf{0.8991} & \textbf{0.1081} & \textbf{0.8808} & \textbf{0.1001} & \textbf{0.9060} \\
\midrule
\multirow{11}{*}{KITTI}
& FP      & 32/32 & 0.0818 & 0.9369 & 0.0804 & 0.9396 & 0.0832 & 0.9340 & 0.0814 & 0.9389 \\
\cline{2-11}
& MinMax \cite{quant_survey} & 4/4 & 0.3770 & 0.3441 & 1.9612 & 0.2058 & 0.3938 & 0.3423 & 4.4358 & 0.0832 \\
& Percentile \cite{Percentile} & 4/4 & 0.3418 & 0.4099 & 0.3876 & 0.3327 & 0.3668 & 0.3780 & 0.3932 & 0.3275 \\
& BRECQ    \cite{BRECQ}    & 4/4   & 0.3719 & 0.3522 & 0.3989 & 0.3160 & 0.3906 & 0.3344 & 0.3990 & 0.3175 \\
& QDrop    \cite{QDROP}    & 4/4   & 0.3934 & 0.3234 & 0.3855 & 0.3338 & 0.3620 & 0.3748 & 0.3412 & 0.4082 \\
& PTQ4ViT \cite{ptq4vit} & 4/4 & 0.3439 & 0.4106 & 0.3923 & 0.3251 & 0.3308 & 0.4187 & 0.3972 & 0.3200 \\
& RepQ-ViT \cite{RepQ-ViT} & 4/4   & 0.3434 & 0.4159 & 0.2539 & 0.5410 & 0.3634 & 0.3897 & 0.3906 & 0.3287 \\
& ERQ \cite{erq} & 4/4 & 0.3178 & 0.4847 & 0.3528 & 0.4241 & 0.3176 & 0.4616 & 0.3490 & 0.4066 \\
& QwT \cite{qwt} & 4/4 & 0.1802 & 0.6862 & 0.2417 & 0.5867 & \underline{0.1951} & \underline{0.6941} & 0.2539 & 0.5346\\
& QSCA     \cite{QSCA}     & 4/4   & \underline{0.1874} & \underline{0.7273} & \underline{0.2365} & \underline{0.6203} & 0.2067 & 0.6794 & \underline{0.2296} & \underline{0.6174} \\ 
& \textbf{SPA-Q (Ours)} & 4/4 & \textbf{0.1622} & \textbf{0.7826} & \textbf{0.1840} & \textbf{0.7230} & \textbf{0.1807} & \textbf{0.7449} & \textbf{0.1696} & \textbf{0.7702} \\

\Xhline{1.4pt}

\end{tabular}}
\label{tab:appendix_full_comparison}
\end{table}

Table~\ref{tab:appendix_full_comparison} provides an extended comparison with additional PTQ baselines on NYUv2 \cite{nyu} and KITTI \cite{kitti} under 4-bit quantization.
In addition to the baselines reported in the main paper, we include MinMax \cite{quant_survey}, Percentile \cite{Percentile}, PTQ4ViT \cite{ptq4vit}, ERQ \cite{erq}, and QwT \cite{qwt} to further evaluate the robustness of SPA-Q across different calibration and reconstruction strategies.
Across Depth Anything v1~\cite{DAv1} and v2~\cite{DAv2} with both ViT-S and ViT-B backbones, SPA-Q achieves the best performance in all evaluated settings. 
On NYUv2, SPA-Q consistently improves both AbsRel and \(\delta_1\) over all compared methods.
The improvement is more pronounced for ViT-B backbones, where preserving attention distributions and aligning intermediate feature distributions are especially important under low-bit quantization.
On KITTI, SPA-Q also achieves the best results across all model variants, showing that the proposed framework remains effective in outdoor driving scenes. 
Although QwT provides a strong baseline in some KITTI settings, SPA-Q still achieves lower AbsRel and higher \(\delta_1\), demonstrating the benefit of explicitly preserving depth-relevant intermediate representations.

Overall, these extended results confirm that the gains of SPA-Q are not limited to the main PTQ baselines.
Instead, SPA-Q remains consistently effective against a broader set of PTQ methods, further supporting the importance of attention-preserving calibration and channel-wise distribution alignment for quantized monocular depth estimation.

\newpage

\begin{table}[H]
\centering
\caption{Extended comparison with additional PTQ baselines on NYUv2 \cite{nyu} and KITTI \cite{kitti} for zero-shot relative depth estimation under W4A6 and W4A8 quantization.
\textit{E.} denotes the encoder backbone used in the MDE architectures.}

\setlength{\tabcolsep}{10pt}
\renewcommand{\arraystretch}{1.5}
\resizebox{\textwidth}{!}{
\begin{tabular}{c l c c c c c c c c c}
\Xhline{1.4pt}
\multirow{3}{*}{\textbf{Dataset}} & \multirow{3}{*}{\textbf{Method}} & \multirow{3}{*}{\textbf{W/A}}
& \multicolumn{4}{c}{\textbf{Depth Anything v1}} 
& \multicolumn{4}{c}{\textbf{Depth Anything v2}} \\
\cmidrule(lr){4-7} \cmidrule(lr){8-11}
& & 
& \multicolumn{2}{c}{\textbf{\textit{E.} ViT-S}} 
& \multicolumn{2}{c}{\textbf{\textit{E.} ViT-B}} 
& \multicolumn{2}{c}{\textbf{\textit{E.} ViT-S}} 
& \multicolumn{2}{c}{\textbf{\textit{E.} ViT-B}} \\
& & 
& AbsRel $\downarrow$ & $\delta_1 \uparrow$
& AbsRel $\downarrow$ & $\delta_1 \uparrow$
& AbsRel $\downarrow$ & $\delta_1 \uparrow$
& AbsRel $\downarrow$ & $\delta_1 \uparrow$ \\
\midrule
\multirow{21}{*}{NYUv2}
& FP 
& 32/32 
& 0.0525 & 0.9720 & 0.0459 & 0.9791 
& 0.0513 & 0.9736 & 0.0460 & 0.9770 \\
\cline{2-11}
& MinMax \cite{quant_survey}     & 4/6 
& 0.2417 & 0.5632 
& 0.2748 & 0.4973 
& 0.2599 & 0.5324 
& 0.2866 & 0.4738 \\

& Percentile \cite{Percentile} & 4/6 
& 0.1050 & 0.8837 
& 0.0958 & 0.9071 
& 0.0891 & 0.9196 
& 0.0831 & 0.9355 \\

& BRECQ \cite{BRECQ}      & 4/6 
& 0.2337 & 0.5786 
& 0.2462 & 0.5542 
& 0.2611 & 0.5269 
& 0.2697 & 0.5084 \\

& QDrop \cite{QDROP}     & 4/6 
& 0.2071 & 0.6369 
& 0.1823 & 0.6987 
& 0.1700 & 0.7285 
& 0.1560 & 0.7619 \\

& PTQ4ViT \cite{ptq4vit}    & 4/6 
& 0.2166 & 0.6187 
& 0.2201 & 0.6122 
& 0.2642 & 0.5159 
& 0.2275 & 0.6095 \\

& RepQ-ViT \cite{RepQ-ViT}   & 4/6 
& 0.1304 & 0.8256 
& 0.1058 & 0.8867 
& 0.1131 & 0.8657 
& 0.0918 & 0.9158 \\

& ERQ \cite{erq}       & 4/6 
& 0.0945 & 0.9075 
& 0.0766 & 0.9390 
& 0.0874 & 0.9245 
& 0.0914 & 0.9152 \\

& QwT \cite{qwt}        & 4/6 
& 0.0884 & 0.9189 
& 0.0928 & 0.9089 
& 0.0833 & 0.9323 
& 0.1000 & 0.8944 \\

& QSCA \cite{QSCA}       & 4/6
& \underline{0.0810} & \underline{0.9333}
& \underline{0.0739} & \underline{0.9441}
& \underline{0.0757} & \underline{0.9450}
& \underline{0.0726} & \underline{0.9468} \\
& \textbf{SPA-Q (Ours)} 
& 4/6 
& \textbf{0.0732} & \textbf{0.9473} & \textbf{0.0623} & \textbf{0.9597} 
& \textbf{0.0735} & \textbf{0.9485} & \textbf{0.0647} & \textbf{0.9603} \\
\cline{2-11}
& MinMax \cite{quant_survey}     & 4/8 
& 0.1522 & 0.7740 
& 0.2253 & 0.5934 
& 0.1345 & 0.8187 
& 0.2507 & 0.5465 \\

& Percentile \cite{Percentile} & 4/8 
& 0.0861 & 0.9233 
& 0.0648 & 0.9588 
& 0.0798 & 0.9375 
& 0.0604 & 0.9621 \\

& BRECQ \cite{BRECQ}       & 4/8 
& 0.1182 & 0.8533 
& 0.2152 & 0.6166 
& 0.0999 & 0.8927 
& 0.2193 & 0.6095 \\

& QDrop \cite{QDROP}      & 4/8 
& 0.0700 & 0.9504 
& \underline{0.0599} & 0.9627 
& \textbf{0.0651} & \textbf{0.9592} 
& 0.0601 & 0.9621 \\

& PTQ4ViT \cite{ptq4vit}    & 4/8 
& 0.2001 & 0.6544 
& 0.1877 & 0.6837 
& 0.2596 & 0.5246 
& 0.2190 & 0.6328 \\

& RepQ-ViT \cite{RepQ-ViT}   & 4/8 
& 0.1066 & 0.8833 
& 0.0864 & 0.9259 
& 0.1066 & 0.8792 
& 0.0788 & 0.9367 \\

& ERQ \cite{erq}        & 4/8 
& 0.0812 & 0.9333 
& 0.0608 & \underline{0.9632} 
& 0.0809 & 0.9355 
& \underline{0.0597} & \underline{0.9641} \\

& QwT  \cite{qwt}       & 4/8 
& 0.0769 & 0.9393 
& 0.0789 & 0.9348 
& 0.0663 & 0.9564 
& 0.0808 & 0.9300 \\

& QSCA \cite{QSCA}       & 4/8
& \underline{0.0682} & \underline{0.9551}
& 0.0657 & 0.9542
& 0.0679 & 0.9549
& 0.0599 & 0.9635 \\
& \textbf{SPA-Q (Ours)} 
& 4/8 
& \textbf{0.0664} & \textbf{0.9574} & \textbf{0.0560} & \textbf{0.9672} 
& \textbf{0.0651} & \underline{0.9591} & \textbf{0.0562} & \textbf{0.9682} \\
\midrule
\multirow{21}{*}{KITTI}
& FP 
& 32/32 
& 0.0818 & 0.9369 & 0.0804 & 0.9396 
& 0.0832 & 0.9340 & 0.0814 & 0.9389 \\
\cline{2-11}
& MinMax \cite{quant_survey}     & 4/6 & 0.3290 & 0.4467 & 0.3769 & 0.3586 & 0.3744 & 0.3609 & 0.3982 & 0.3161 \\
& Percentile \cite{Percentile} & 4/6 & 0.1269 & 0.8558 & 0.1687 & 0.7580 & 0.1283 & 0.8613 & 0.1866 & 0.7219 \\
& BRECQ \cite{BRECQ}       & 4/6 & 0.2851 & 0.5223 & 0.3529 & 0.4035 & 0.3332 & 0.4320 & 0.3497 & 0.4093 \\
& QDrop \cite{QDROP}      & 4/6 & 0.2408 & 0.6052 & 0.2611 & 0.5897 & 0.2790 & 0.5351 & 0.3037 & 0.5268 \\
& PTQ4ViT \cite{ptq4vit}    & 4/6 & 0.3230 & 0.4446 & 0.3848 & 0.3488 & 0.3614 & 0.3931 & 0.3917 & 0.3308 \\
& RepQ-ViT \cite{RepQ-ViT}   & 4/6 & 0.1502 & 0.7888 & 0.2602 & 0.5610 & 0.1292 & 0.8293 & 0.1458 & 0.8109 \\
& ERQ \cite{erq}        & 4/6
            & 0.1244 & 0.8629
            & \underline{0.1133} & 0.8705
            & \underline{0.1144} & 0.8770
            & 0.1115 & 0.8685 \\
& QwT \cite{qwt}        & 4/6
            & \textbf{0.1152} & 0.8844
            & 0.1759 & 0.7292
            & 0.1169 & \underline{0.8872}
            & 0.1550 & 0.7700 \\
& QSCA  \cite{QSCA}      & 4/6
            & \underline{0.1161} & \textbf{0.8857}
            & 0.1174 & \underline{0.8722}
            & \textbf{0.1124} & \textbf{0.8893}
            & \underline{0.1067} & \underline{0.8873} \\
& \textbf{SPA-Q (Ours)} 
& 4/6 
& 0.1247 & \underline{0.8849} & \textbf{0.1023} & \textbf{0.8907} 
& 0.1684 & 0.7892 & \textbf{0.0994} & \textbf{0.9003} \\
\cline{2-11}
& MinMax \cite{quant_survey}     & 4/8 & 0.3486 & 0.4215 & 0.3275 & 0.4845 & 0.3529 & 0.4237 & 0.3327 & 0.4376 \\
& Percentile \cite{Percentile} & 4/8 & 0.1111 & 0.8740 & 0.1263 & 0.8564 & 0.1051 & 0.8892 & 0.1045 & 0.9003 \\
& BRECQ \cite{BRECQ}       & 4/8 & 0.3532 & 0.4070 & 0.3317 & 0.4513 & 0.3194 & 0.4575 & 0.3239 & 0.4493 \\
& QDrop \cite{QDROP}      & 4/8 & 0.3071 & 0.4883 & 0.2754 & 0.5422 & 0.2713 & 0.5368 & 0.2688 & 0.5616 \\
& PTQ4ViT \cite{ptq4vit}    & 4/8 & 0.3224 & 0.4580 & 0.3813 & 0.3550 & 0.3538 & 0.3909 & 0.3969 & 0.3205 \\
& RepQ-ViT \cite{RepQ-ViT}   & 4/8 & 0.1375 & 0.8145 & 0.2227 & 0.6436 & 0.1212 & 0.8431 & 0.1136 & 0.8683 \\
& ERQ \cite{erq}        & 4/8
            & 0.1164 & 0.8801
            & \underline{0.1005} & 0.8937
            & \underline{0.1040} & 0.8881
            & 0.0988 & 0.9043 \\
& QwT \cite{qwt}        & 4/8
            & \underline{0.1044} & 0.8970
            & 0.1698 & 0.7362
            & 0.1061 & \textbf{0.9017}
            & 0.1328 & 0.8228 \\
& QSCA \cite{QSCA}       & 4/8
            & \textbf{0.1019} & \underline{0.9022}
            & 0.1045 & \underline{0.8963}
            & \textbf{0.1020} & \underline{0.9010}
            & \underline{0.0931} & \textbf{0.9161} \\
& \textbf{SPA-Q (Ours)} 
& 4/8 
& 0.1072 & \textbf{0.9035} & \textbf{0.0904} & \textbf{0.9148}
& 0.1057 & 0.8990 & \textbf{0.0920} & \underline{0.9135} \\
\Xhline{1.4pt}
\end{tabular}}
\label{tab:appendix_full_comparison_w4a6_w4a8}
\end{table}

We further evaluate SPA-Q on additional MDE models, MiDaS-BEiT-B-384 \cite{MiDaS}, DPT-L-384 \cite{DPT}, and ZoeDepth \cite{zoedepth}, as shown in Table~\ref{tab:appendix_midas_dpt}.
Models with BEiT backbones, namely MiDaS and ZoeDepth, suffer more severe degradation under 4-bit quantization than DPT-L.
Nevertheless, SPA-Q consistently outperforms the state-of-the-art QSCA \cite{QSCA} across all models and datasets.

\newpage

\section{Generalization Across MDE Models}
\begin{table}[H]
\centering
\caption{Comparison of PTQ methods on NYUv2 \cite{nyu} and KITTI \cite{kitti} with additional MDE architectures, MiDaS-BEiT-B-384 \cite{MiDaS}, DPT-L-384 \cite{DPT}, and ZoeDepth \cite{zoedepth}, under 4-bit quantization.}
\setlength{\tabcolsep}{10pt}
\renewcommand{\arraystretch}{1.3}
\footnotesize
\resizebox{\textwidth}{!}{
\begin{tabular}{c l c c c c c c c}
\Xhline{1.4pt}
\multirow{2}{*}{\textbf{Dataset}} & \multirow{2}{*}{\textbf{Method}} & \multirow{2}{*}{\textbf{W/A}}
& \multicolumn{2}{c}{\textbf{MiDaS-BEiT-B-384}} 
& \multicolumn{2}{c}{\textbf{DPT-L-384}} 
& \multicolumn{2}{c}{\textbf{ZoeDepth (Metric)}} \\
\cmidrule(lr){4-5} \cmidrule(lr){6-7} \cmidrule(lr){8-9}
& & 
& AbsRel $\downarrow$ & $\delta_1 \uparrow$
& AbsRel $\downarrow$ & $\delta_1 \uparrow$
& AbsRel $\downarrow$ & $\delta_1 \uparrow$ \\
\midrule
\multirow{4}{*}{NYUv2}
& FP 
& 32/32 
& 0.0663 & 0.9568 
& 0.0947 & 0.9112 
& 0.0755 & 0.9544 \\
\cline{2-9}
& RepQ-ViT \cite{RepQ-ViT} 
& 4/4 
& 0.2603 & 0.5254 
& 0.1394 & 0.8139 
& 0.4949 & 0.1033 \\
& QSCA \cite{QSCA} 
& 4/4 
& \underline{0.2253} & \underline{0.5998} 
& \underline{0.1380} & \underline{0.8192} 
& \underline{0.3969} & \underline{0.2054} \\
& \textbf{SPA-Q (Ours)} 
& 4/4 
& \textbf{0.1803} & \textbf{0.7081} 
& \textbf{0.1276} & \textbf{0.8425} 
& \textbf{0.3051} & \textbf{0.3894} \\
\midrule
\multirow{4}{*}{KITTI}
& FP 
& 32/32 
& 0.1928 & 0.7076 
& 0.1079 & 0.8858 
& 0.0589 & 0.9664 \\
\cline{2-9}
& RepQ-ViT \cite{RepQ-ViT} 
& 4/4 
& 0.3761 & 0.3580 
& 0.1661 & \underline{0.7824} 
& 0.4650 & 0.1529 \\
& QSCA \cite{QSCA} 
& 4/4 
& \underline{0.3740} & \underline{0.3628} 
& \underline{0.1650} & 0.7810 
& \underline{0.2701} & \underline{0.5237} \\
& \textbf{SPA-Q (Ours)} 
& 4/4 
& \textbf{0.3250} & \textbf{0.4383} 
& \textbf{0.1427} & \textbf{0.8379} 
& \textbf{0.2551} & \textbf{0.6171} \\
\Xhline{1.4pt}
\end{tabular}}
\label{tab:appendix_midas_dpt}
\end{table}

To examine whether SPA-Q generalizes beyond Depth Anything, we further evaluate it on three additional MDE models with different backbones and prediction targets: MiDaS-BEiT-B-384 \cite{MiDaS} and DPT-L-384 \cite{DPT} for relative depth estimation, and ZoeDepth \cite{zoedepth} for metric depth estimation.
As shown in Table~\ref{tab:appendix_midas_dpt}, SPA-Q achieves the best AbsRel and $\delta_1$ for all three models on both NYUv2 \cite{nyu} and KITTI \cite{kitti} under 4-bit quantization, consistently outperforming RepQ-ViT \cite{RepQ-ViT} and QSCA \cite{QSCA}.
These results indicate that SPA-Q is not tailored to a specific architecture and is applicable to a broad range of MDE models, including those for metric depth estimation.

\section{Evaluation on Structural Metrics}

\begin{table}[H]
\centering
\caption{Comparison of edge, boundary, and planarity metrics for Depth Anything v1 \cite{DAv1} under W4A4 and W4A6 quantization. Edge F1 \cite{EdgeF1}, Boundary F1 \cite{BoundaryF1}, and ORD \cite{ORD} are evaluated on NYUv2 \cite{nyu}, and planarity errors (PE) are evaluated on iBims-1 \cite{iBims-1}.}
\setlength{\tabcolsep}{10pt}
\renewcommand{\arraystretch}{1.3}
\footnotesize
\resizebox{\textwidth}{!}{
\begin{tabular}{c l c c c c c c}
\Xhline{1.4pt}
\multirow{3}{*}{\textbf{Encoder}} & \multirow{3}{*}{\textbf{Method}} & \multirow{3}{*}{\textbf{W/A}}
& \multicolumn{5}{c}{\textbf{Depth Anything v1}} \\
\cmidrule(lr){4-8}
& & 
& \multicolumn{3}{c}{Boundary \& Ordinal Accuracy} & \multicolumn{2}{c}{Planarity Errors (cm / $^{\circ}$)} \\
\cmidrule(lr){4-6} \cmidrule(lr){7-8}
& & 
& Edge F1@0.5 $\uparrow$ & Boundary F1 $\uparrow$ & ORD $\downarrow$ 
& $\varepsilon_{\mathrm{PE}}^{\mathrm{plan}} \downarrow$ & $\varepsilon_{\mathrm{PE}}^{\mathrm{orie}} \downarrow$ \\
\midrule
\multirow{7}{*}{ViT-S}
& FP 
& 32/32 
& 0.1862 & 0.0438 & 0.1177 
& 1.3337 & 4.4868 \\
\cline{2-8}
& RepQ-ViT \cite{RepQ-ViT} 
& 4/4 
& 0.1025 & 0.0018 & 0.5405 
& \underline{11.4530} & 35.0064 \\
& QSCA \cite{QSCA} 
& 4/4 
& \textbf{0.1292} & \underline{0.0126} & \underline{0.2956} 
& \textbf{10.5740} & \underline{24.1218} \\
& \textbf{SPA-Q (Ours)} 
& 4/4 
& \underline{0.1286} & \textbf{0.0161} & \textbf{0.2423} 
& 13.1100 & \textbf{19.8219} \\
\cline{2-8}
& RepQ-ViT \cite{RepQ-ViT} 
& 4/6 
& 0.1251 & 0.0174 & \underline{0.2795} 
& 5.4846 & 14.5692 \\
& QSCA \cite{QSCA} 
& 4/6 
& \textbf{0.1915} & \textbf{0.0475} & \textbf{0.1627} 
& \textbf{4.0663} & \underline{9.8937} \\
& \textbf{SPA-Q (Ours)} 
& 4/6 
& \underline{0.1907} & \underline{0.0436} & \textbf{0.1627} 
& \underline{4.2679} & \textbf{9.0219} \\
\midrule
\multirow{7}{*}{ViT-B}
& FP 
& 32/32 
& 0.1991 & 0.0477 & 0.1005 
& 1.0963 & 4.1992 \\
\cline{2-8}
& RepQ-ViT \cite{RepQ-ViT} 
& 4/4 
& 0.1011 & 0.0011 & 0.7412 
& 14.1688 & 44.1632 \\
& QSCA \cite{QSCA} 
& 4/4 
& \underline{0.1069} & \underline{0.0018} & \underline{0.4732} 
& \underline{14.0001} & \underline{34.3514} \\
& \textbf{SPA-Q (Ours)} 
& 4/4 
& \textbf{0.1111} & \textbf{0.0091} & \textbf{0.2441} 
& \textbf{13.6755} & \textbf{20.1182} \\
\cline{2-8}
& RepQ-ViT \cite{RepQ-ViT} 
& 4/6 
& 0.1172 & 0.0180 & 0.2418 
& \underline{4.1241} & 15.4422 \\
& QSCA \cite{QSCA} 
& 4/6 
& \underline{0.1658} & \underline{0.0328} & \underline{0.1672} 
& 5.6829 & \underline{11.1425} \\
& \textbf{SPA-Q (Ours)} 
& 4/6 
& \textbf{0.1938} & \textbf{0.0404} & \textbf{0.1377} 
& \textbf{3.7100} & \textbf{7.7550} \\
\Xhline{1.4pt}
\end{tabular}}
\label{tab:appendix_structure_metrics}
\end{table}

We further report structural metrics to evaluate how well each method preserves the geometric structure of depth maps, including boundary and ordinal accuracy metrics (Edge F1, Boundary F1, and ORD) and planarity errors, as shown in Table~\ref{tab:appendix_structure_metrics}.
SPA-Q achieves the best performance on most metrics under both W4A4 and W4A6 quantization, indicating that it effectively preserves fine depth structures such as object boundaries, relative depth ordering, and planar surfaces.

\newpage

\section{Block-wise Feature Similarity Analysis}

\begin{figure}[H]
    \centering
    \includegraphics[width=0.98\textwidth]{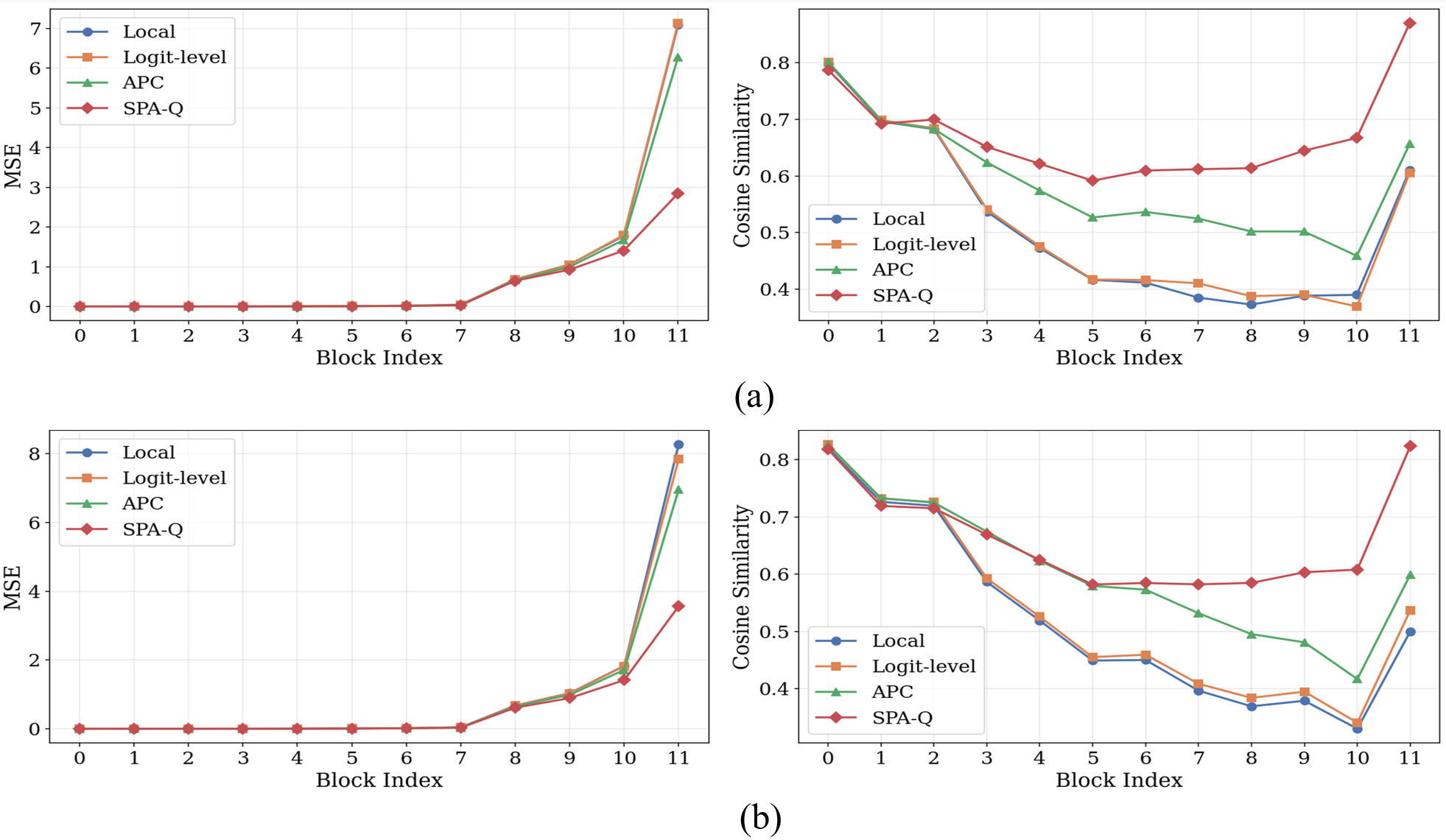}
    \caption{
Block-wise feature similarity analysis under 4-bit quantization on Depth Anything v1.
(a) and (b) show results on NYUv2 \cite{nyu} and KITTI \cite{kitti}.
We compare the MSE and cosine similarity between full-precision and quantized encoder features across transformer blocks.
}
    \label{fig:fig_appendix_blockwise_dav1}
    
\end{figure}

\begin{figure}[H]
    \centering
    \includegraphics[width=0.98\textwidth]{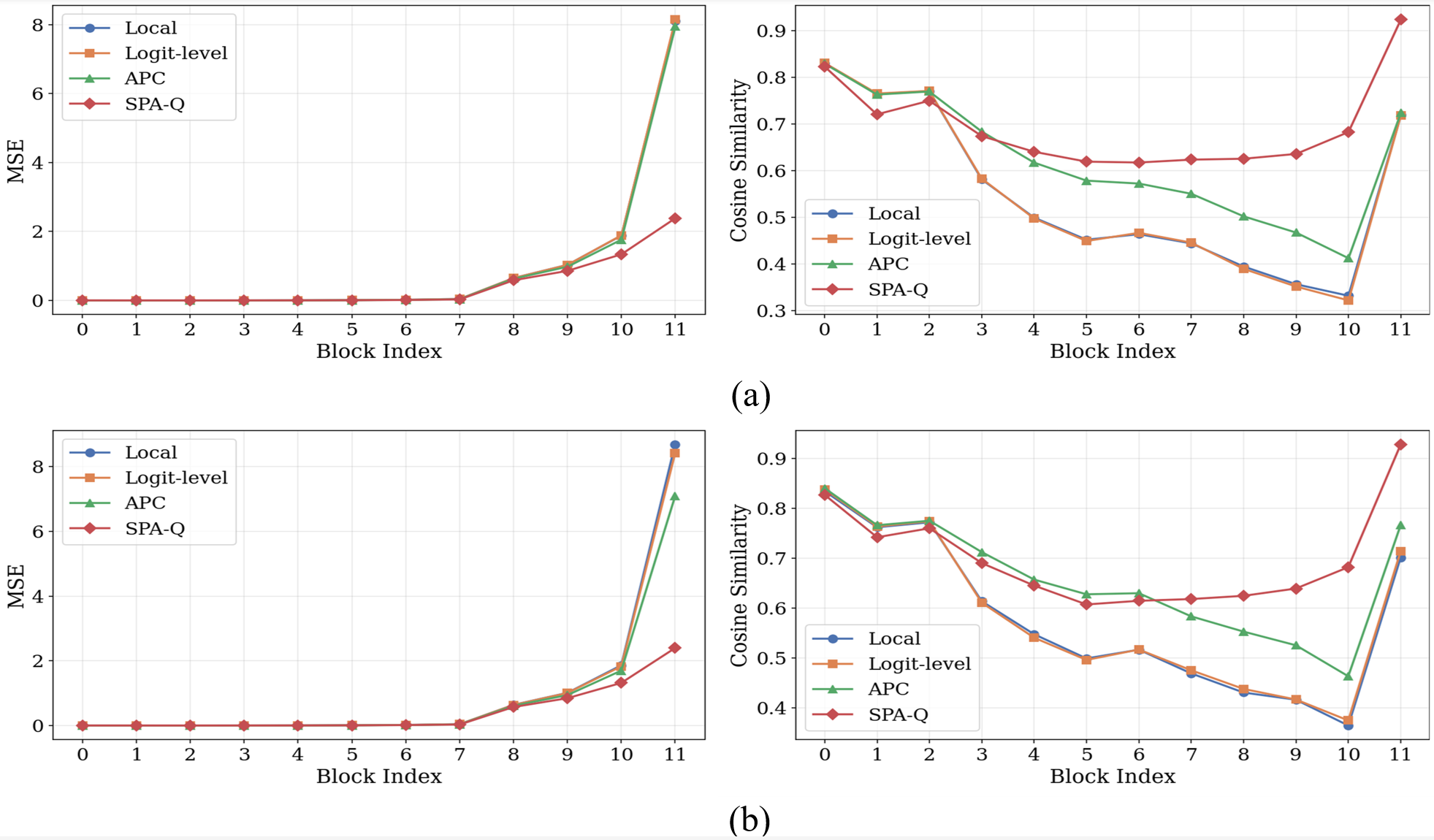}
    \caption{
Block-wise feature similarity analysis under 4-bit quantization on Depth Anything v2.
(a) and (b) show results on NYUv2 \cite{nyu} and KITTI \cite{kitti}.
We compare the MSE and cosine similarity between full-precision and quantized encoder features across transformer blocks.
}
    \label{fig:fig_appendix_blockwise_dav2}
    
\end{figure}

Figure~\ref{fig:fig_appendix_blockwise_dav1} and Figure~\ref{fig:fig_appendix_blockwise_dav2} provide an extended block-wise feature similarity analysis under 4-bit quantization across Depth Anything v1~\cite{DAv1} and v2~\cite{DAv2} on NYUv2 \cite{nyu} and KITTI \cite{kitti}.
For each setting, we compare the MSE and cosine similarity between full-precision and quantized encoder features across transformer blocks.
Across all settings, SPA-Q consistently yields lower feature MSE and higher cosine similarity than the compared calibration objectives, especially in deeper blocks.

\newpage

\section{Qualitative Analysis of Feature Maps}

\begin{figure}[H]
    \centering
    \includegraphics[width=0.98\textwidth]{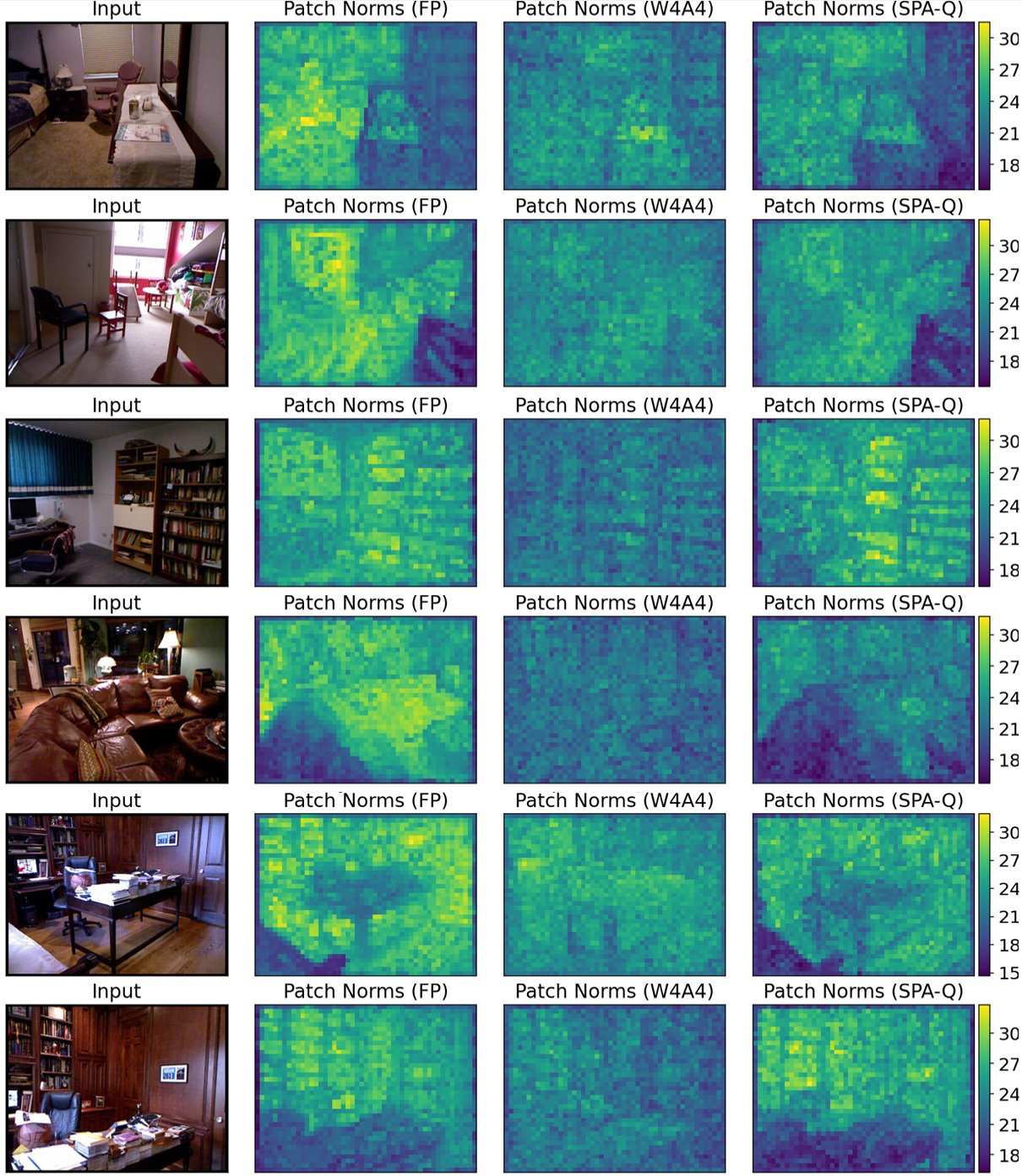}
    \caption{
Feature map visualization for Depth Anything v2 with a ViT-S backbone on NYUv2 \cite{nyu} under 4-bit quantization.
We compare the full-precision model, W4A4 quantization, and SPA-Q.
}
    \label{fig:fig_7_1}
    
\end{figure}

\begin{figure}[H]
    \centering
    \includegraphics[width=0.98\textwidth]{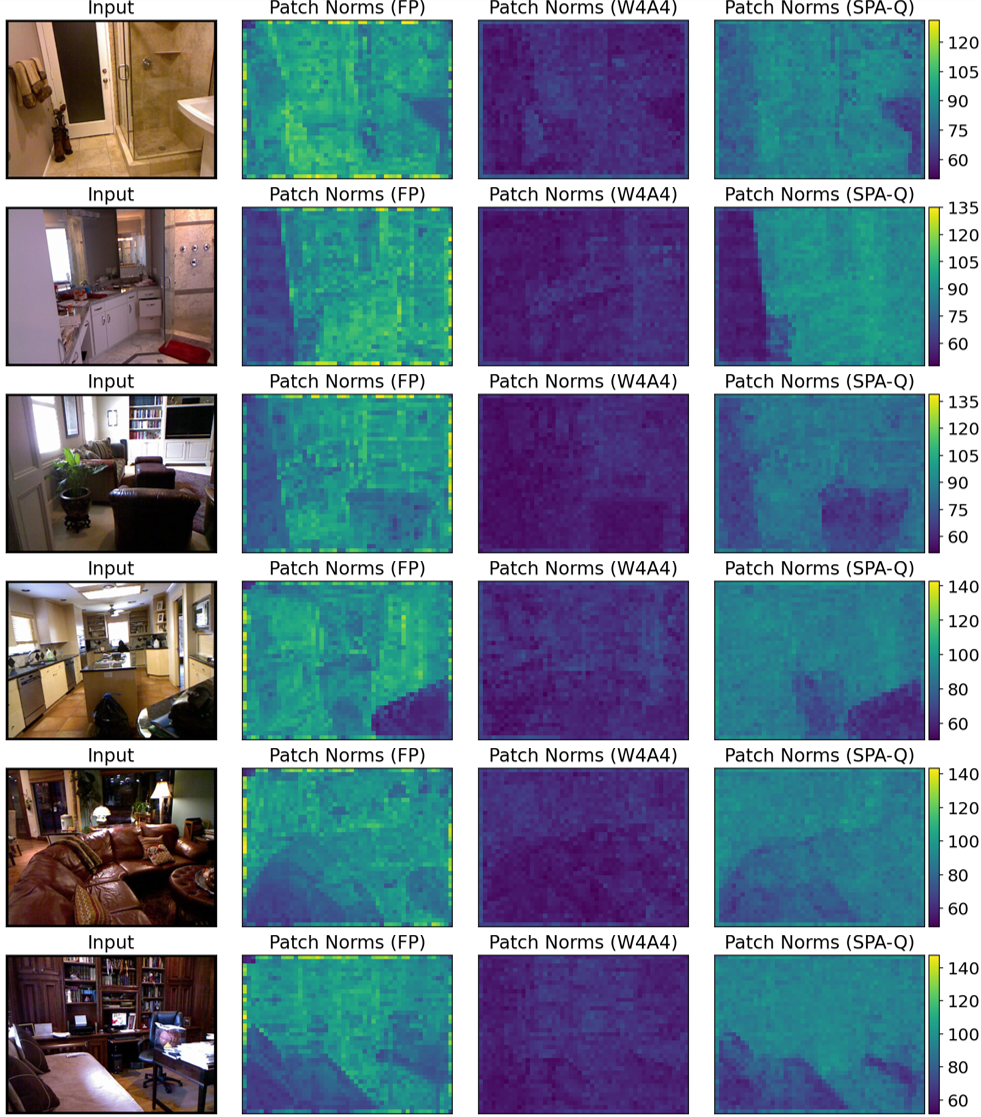}
    \caption{
Feature map visualization for Depth Anything v1 with a ViT-B backbone on NYUv2 \cite{nyu} under 4-bit quantization.
We compare the full-precision model, W4A4 quantization, and SPA-Q.
}
    \label{fig:fig_7_2}
    
\end{figure}

\begin{figure}[H]
    \centering
    \includegraphics[width=0.98\textwidth]{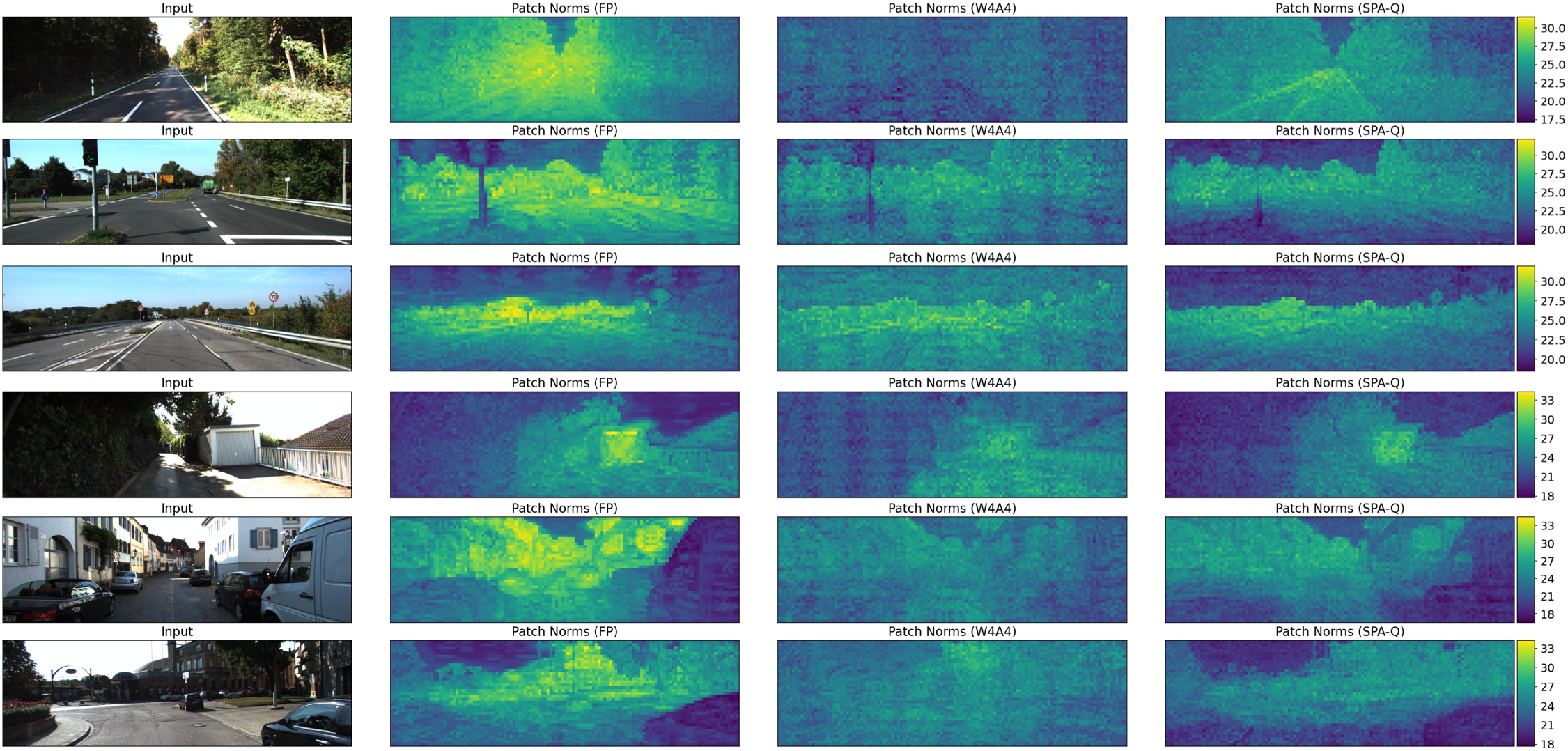}
    \caption{
Feature map visualization for Depth Anything v2 with a ViT-S backbone on KITTI \cite{kitti} under 4-bit quantization.
We compare the full-precision model, W4A4 quantization, and SPA-Q.
}
    \label{fig:fig_8_1}
    
\end{figure}

\begin{figure}[H]
    \centering
    \includegraphics[width=0.98\textwidth]{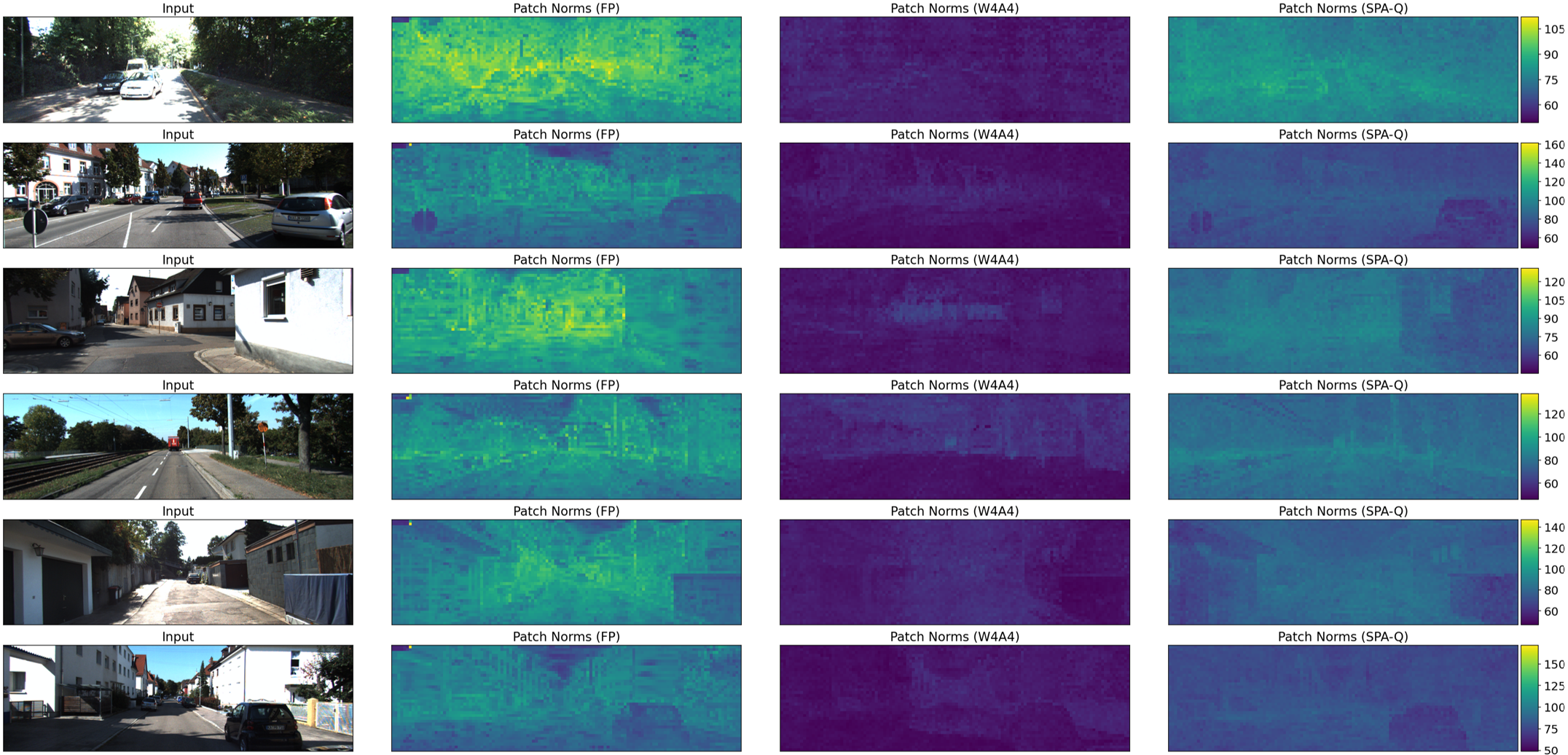}
    \caption{
Feature map visualization for Depth Anything v1 with a ViT-B backbone on KITTI \cite{kitti} under 4-bit quantization.
We compare the full-precision model, W4A4 quantization, and SPA-Q.
}
    \label{fig:fig_8_2}
    
\end{figure}

Figures~\ref{fig:fig_7_1}, ~\ref{fig:fig_7_2} visualize feature maps on NYUv2 \cite{nyu}, while Figures~\ref{fig:fig_8_1} and~\ref{fig:fig_8_2} show the corresponding results on KITTI \cite{kitti}.
We compare the full-precision model, naive W4A4 quantization, and SPA-Q.
Naive W4A4 quantization substantially distorts the spatial activation patterns of intermediate features, which the decoder relies on for dense depth prediction.
In contrast, SPA-Q yields feature maps with higher fidelity to their full-precision counterparts, preserving spatial layouts, object regions, and boundary-related responses.
These results show that SPA-Q mitigates feature distortion caused by low-bit quantization across different Depth Anything models, backbones, and datasets.

\newpage

\newpage

\newpage

\newpage
\section*{NeurIPS Paper Checklist}

\begin{enumerate}

\item {\bf Claims}
    \item[] Question: Do the main claims made in the abstract and introduction accurately reflect the paper's contributions and scope?
    \item[] Answer: \answerYes{} %
    \item[] Justification: The main claims in the abstract and introduction match the paper’s contributions and experimental results. The paper claims that SPA-Q addresses attention-preservation and intermediate feature distribution shift in low-bit PTQ for MDE, and these claims are supported by the method description and experiments in Secs. 1, 3, and 4.

\item {\bf Limitations}
    \item[] Question: Does the paper discuss the limitations of the work performed by the authors?
    \item[] Answer: \answerYes{} %
    \item[] Justification: \item[] Justification: The paper explicitly discusses limitations in the Limitations \& Future Works section. In particular, it notes that a noticeable gap to full precision still remains under severe low-bit quantization, that quantized depth maps can appear less smooth due to the restricted set of representable depth values, and that the current study is limited to monocular depth estimation rather than stereo or video depth estimation.

\item {\bf Theory assumptions and proofs}
    \item[] Question: For each theoretical result, does the paper provide the full set of assumptions and a complete (and correct) proof?
    \item[] Answer: \answerNA{} %
    \item[] Justification: The paper does not include formal theoretical results such as theorems, lemmas, propositions, or proofs. It presents objective formulations and empirical analyses, but no claims requiring formal proof.

\item {\bf Experimental result reproducibility}
    \item[] Question: Does the paper fully disclose all the information needed to reproduce the main experimental results of the paper to the extent that it affects the main claims and/or conclusions of the paper (regardless of whether the code and data are provided or not)?
    \item[] Answer: \answerYes{} %
    \item[] Justification: The main experiments are conducted on publicly available models and datasets, and the paper specifies the model variants, datasets, evaluation metrics, calibration protocol, optimization objectives, and training details needed for reproduction. Since the evaluation is performed on open benchmarks with standard settings, the reported results can be reproduced by other researchers.

\item {\bf Open access to data and code}
    \item[] Question: Does the paper provide open access to the data and code, with sufficient instructions to faithfully reproduce the main experimental results, as described in supplemental material?
    \item[] Answer: \answerNo{} %
    \item[] Justification: The paper evaluates the proposed method on publicly available datasets and provides the experimental details needed for reproduction, but it does not release implementation code, reproduction scripts, or an anonymized repository link in the current submission.

\item {\bf Experimental setting/details}
    \item[] Question: Does the paper specify all the training and test details (e.g., data splits, hyperparameters, how they were chosen, type of optimizer) necessary to understand the results?
    \item[] Answer: \answerYes{} %
    \item[] Justification: The paper specifies the main experimental details needed to understand the reported results, including the evaluated model variants and backbones, datasets and evaluation metrics, calibration set construction, quantization setting, CWDA initialization, optimization strategy, loss weighting, optimizer, learning rate, batch size, and hardware configuration.

\item {\bf Experiment statistical significance}
    \item[] Question: Does the paper report error bars suitably and correctly defined or other appropriate information about the statistical significance of the experiments?
    \item[] Answer: \answerNo{} %
    \item[] Justification: The paper does not report formal measures of statistical significance. However, the experiments follow the standard PTQ evaluation protocol used in prior work, and all methods are compared under the same quantization setting, calibration procedure, and evaluation benchmarks.

\item {\bf Experiments compute resources}
    \item[] Question: For each experiment, does the paper provide sufficient information on the computer resources (type of compute workers, memory, time of execution) needed to reproduce the experiments?
    \item[] Answer: \answerYes{} %
    \item[] Justification: The paper reports that all experiments are conducted on a single RTX 6000 Ada Generation GPU, and also provides model size, parameter count, and BOPs for the evaluated models.
    
\item {\bf Code of ethics}
    \item[] Question: Does the research conducted in the paper conform, in every respect, with the NeurIPS Code of Ethics \url{https://neurips.cc/public/EthicsGuidelines}?
    \item[] Answer: \answerYes{} %
    \item[] Justification: We have reviewed the NeurIPS Code of Ethics, and this work conforms to it. The paper studies model quantization for monocular depth estimation using standard public benchmarks and does not involve human subjects, personal data collection, or high-risk model release.

\item {\bf Broader impacts}
    \item[] Question: Does the paper discuss both potential positive societal impacts and negative societal impacts of the work performed?
    \item[] Answer: \answerNo{} %
    \item[] Justification:
    The manuscript does not include a dedicated broader impacts discussion because this work focuses on a PTQ method for existing MDE models, rather than introducing a new dataset, deployed system, or application. Its societal impacts are mainly indirect and tied to the underlying MDE applications, while the paper focuses on model compression and evaluation under standard benchmarks.
    
\item {\bf Safeguards}
    \item[] Question: Does the paper describe safeguards that have been put in place for responsible release of data or models that have a high risk for misuse (e.g., pre-trained language models, image generators, or scraped datasets)?
    \item[] Answer: \answerNA{} %
    \item[] Justification: This paper does not release new datasets, models, or other assets that require special safeguards for responsible use. Therefore, this question is not applicable to the current work.
    \begin{itemize}
        \item The answer \answerNA{} means that the paper poses no such risks.
        \item Released models that have a high risk for misuse or dual-use should be released with necessary safeguards to allow for controlled use of the model, for example by requiring that users adhere to usage guidelines or restrictions to access the model or implementing safety filters. 
        \item Datasets that have been scraped from the Internet could pose safety risks. The authors should describe how they avoided releasing unsafe images.
        \item We recognize that providing effective safeguards is challenging, and many papers do not require this, but we encourage authors to take this into account and make a best faith effort.
    \end{itemize}

\item {\bf Licenses for existing assets}
    \item[] Question: Are the creators or original owners of assets (e.g., code, data, models), used in the paper, properly credited and are the license and terms of use explicitly mentioned and properly respected?
    \item[] Answer: \answerYes{} %
    \item[] Justification:
    The paper uses publicly available datasets, models, and prior methods, and properly credits their creators through citations. We respect the corresponding licenses and terms of use for all existing assets used in this work.

\item {\bf New assets}
    \item[] Question: Are new assets introduced in the paper well documented and is the documentation provided alongside the assets?
    \item[] Answer: \answerNA{} %
    \item[] Justification: The paper does not introduce or release new datasets, code, models, or other assets. Therefore, this question is not applicable to the current work.

\item {\bf Crowdsourcing and research with human subjects}
    \item[] Question: For crowdsourcing experiments and research with human subjects, does the paper include the full text of instructions given to participants and screenshots, if applicable, as well as details about compensation (if any)? 
    \item[] Answer: \answerNA{} %
    \item[] Justification: The paper does not involve crowdsourcing or research with human subjects. Therefore, this question is not applicable to the current work.

\item {\bf Institutional review board (IRB) approvals or equivalent for research with human subjects}
    \item[] Question: Does the paper describe potential risks incurred by study participants, whether such risks were disclosed to the subjects, and whether Institutional Review Board (IRB) approvals (or an equivalent approval/review based on the requirements of your country or institution) were obtained?
    \item[] Answer: \answerNA{} %
    \item[] Justification: The paper does not involve human subjects research or crowdsourcing. Therefore, IRB approval or an equivalent review is not applicable to this work.

\item {\bf Declaration of LLM usage}
    \item[] Question: Does the paper describe the usage of LLMs if it is an important, original, or non-standard component of the core methods in this research? Note that if the LLM is used only for writing, editing, or formatting purposes and does \emph{not} impact the core methodology, scientific rigor, or originality of the research, declaration is not required.
    \item[] Answer: \answerNA{} %
    \item[] Justification: Large language models were used only for writing assistance in preparing the manuscript. They were not used in the core methodology, experiments, or analysis, and therefore did not affect the scientific content or conclusions of the paper.

\end{enumerate}

\end{document}